\documentclass{article}

\usepackage{microtype}
\usepackage{graphicx}
\usepackage{subfigure}
\usepackage{multirow}
\usepackage{booktabs} 

\usepackage{hyperref}

\usepackage[accepted]{icml2024}

\usepackage{amsmath}
\usepackage{amssymb}
\usepackage{mathtools}
\usepackage{amsthm}

\usepackage[capitalize,noabbrev]{cleveref}

\theoremstyle{plain}
\newtheorem{theorem}{Theorem}[section]

\newtheorem{lemma}[theorem]{Lemma}
\newtheorem{ansatz}[theorem]{Ansatz}

\theoremstyle{definition}

\theoremstyle{remark}

\DeclareMathOperator*{\argmin}{argmin}  
\usepackage[textsize=tiny]{todonotes}

\icmltitlerunning{Characteristic Guidance: Non-linear Correction for Diffusion Model at Large Guidance Scale}

\begin{document}

\twocolumn[
\icmltitle{Characteristic Guidance:\\ Non-linear Correction for Diffusion Model at Large Guidance Scale}



\icmlsetsymbol{equal}{*}

\begin{icmlauthorlist}
\icmlauthor{Candi Zheng}{equal,yyy,sch}
\icmlauthor{Yuan Lan}{equal}
\end{icmlauthorlist}

\icmlaffiliation{yyy}{Department of Mathematics, Hong Kong University of Science and Technology, Hong Kong SAR, China}
\icmlaffiliation{sch}{Department of Mechanics and Aerospace Engineering, Southern University of Science and Technology, Shenzhen, China}

\icmlcorrespondingauthor{Candi Zheng}{czhengac@connect.ust.hk}

\icmlkeywords{Machine Learning, ICML}

\vskip 0.3in
]



\printAffiliationsAndNotice{\icmlEqualContribution} 

\begin{abstract}
Popular guidance for denoising diffusion probabilistic model (DDPM) linearly combines distinct conditional models together to provide enhanced control over samples. However, this approach overlooks nonlinear effects that become significant when guidance scale is large. To address this issue, we propose characteristic guidance, a guidance method that provides first-principle non-linear correction for classifier-free guidance. Such correction forces the guided DDPMs to respect the Fokker-Planck (FP) equation of diffusion process, in a way that is training-free and compatible with existing sampling methods. Experiments show that characteristic guidance enhances semantic characteristics of prompts and mitigate irregularities in image generation, proving effective in diverse applications ranging from simulating magnet phase transitions to latent space sampling.

\end{abstract}

\section{Introduction}
\label{Introduction}

\begin{figure*}[h]
\begin{center}
\includegraphics[width=2\columnwidth]{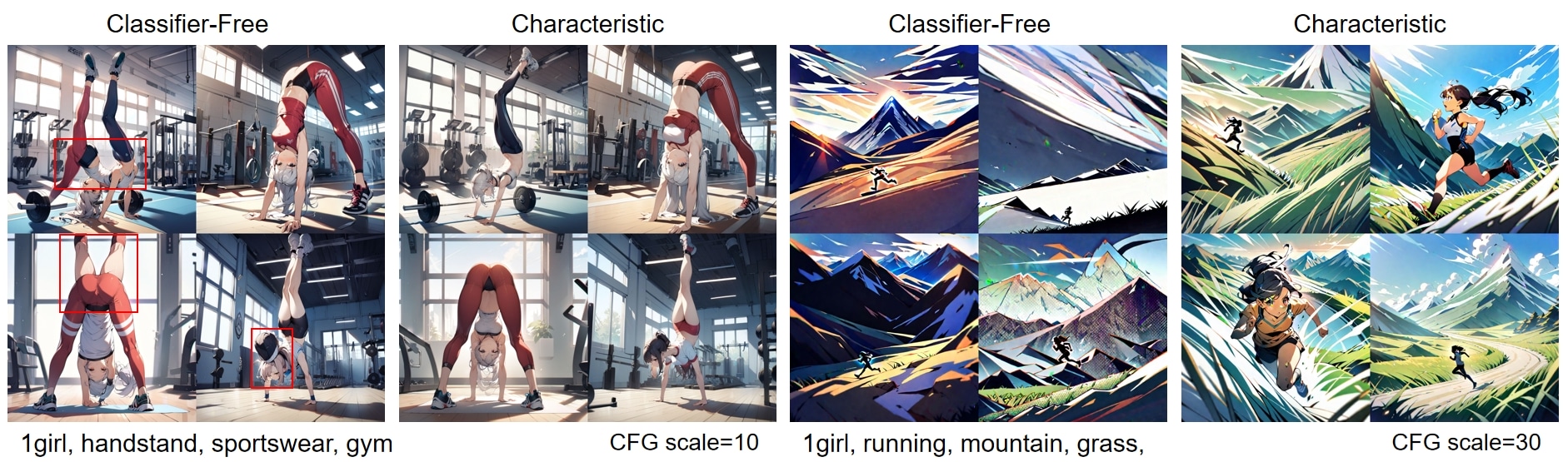}
\end{center}
\vspace{-0.5cm}
\caption{Comparative visualization of images sampled from Stable diffusion XL \cite{podell2023sdxl} between Classifier Free Guidance and Characteristic Guidance (Model name: animagineXL 3.0 \cite{animagine_xl_3_0}, Seeds: 0,1,2,3). By addressing the non-linear effects of the FP equation, characteristic guidance demonstrates ability to mitigate irregularity in color, exposure and anatomy and enhancing prompt's semantic characteristics (e.g., ``grass" depicted in the right-hand example).} \label{SdXL}
\end{figure*}

The diffusion model \citep{SohlDickstein2015DeepUL, Song2019GenerativeMB, Song2020ScoreBasedGM} is a family of generative models that produce high-quality samples utilizing the diffusion processes of molecules. Denoising diffusion probabilistic model (DDPM) \citep{Ho2020DenoisingDP} is one of the most popular diffusion models whose diffusion process can be viewed as a sequence of denoising steps. It is the core of several large text-to-image models that have been widely adopted in AI art production, such as the latent diffusion model (stable diffusion) \citep{Rombach2021HighResolutionIS} and DALL·E \citep{betker2023improving}. DDPMs control the generation of samples by learning conditional distributions that are specified by sample contexts. However, such conditional DDPMs has weak control over context hence do not work well to generate desired content \citep{Luo2022UnderstandingDM}.

Guidance techniques, notably classifier guidance \citep{Song2020ScoreBasedGM,Dhariwal2021DiffusionMB} and classifier-free guidance \citep{Ho2022ClassifierFreeDG}, provide enhanced control at the cost of sample diversity. Classifier guidance, requiring an additional classifier, faces implementation challenges in non-classification tasks like text-to-image generation. Classifier-free guidance circumvents this by linearly combining conditional and unconditional DDPM weighted by a guidance scale parameter. However, large guidance scale often leads to overly saturated and unnatural images \cite{Ho2022ClassifierFreeDG}. Though techniques like dynamic thresholding \cite{saharia2022photorealistic} can handle color issues by clipping out-of-range pixel values, a systematic solution for general sampling tasks, including latent space diffusion \citep{Rombach2021HighResolutionIS} and those beyond image generation, remains absent.

This paper proposes characteristic guidance as a systematic solution to the large guidance scale issue by addressing the non-linearity neglicted in classifier-free guidance. The name \textit{characteristic} serves a dual purpose: Technically, characteristic guidance utilizes \textit{the method of characteristics} in solving the score Fokker-Planck (FP) equation \cite{Lai2022FPDiffusionIS} for diffusion process, addressing its non-linearity to which the classifier-free guidance fails to adhere. It demonstrates clear theoretical advantages on Gaussian models in removing large guidance scale irregularities. 
Experimentally, characteristic guidance notably enhances the \textit{semantic characteristics} of images, ensuring closer alignment with specified conditions. This is evident in our image generation experiments on CIFAR-10, ImageNet 256, and Stable diffusion (Fig.\ref{SdXL}). Notably, it also mitigates color and exposure issues common at large-scale guidance. Furthermore, characteristic guidance is compatible with any continuous data type and is robust across continuous diverse applications, ranging from simulating magnet phase transitions to latent space sampling.

\section{Related Works}
\label{Related Works}

\subsection{Diffusion Models and Guidance Techniques} 
Diffusion models, notably Denoising Diffusion Probabilistic Models (DDPMs) \citep{Ho2020DenoisingDP}, have become pivotal in generating high-quality AI art \citep{SohlDickstein2015DeepUL, Song2019GenerativeMB, Song2020ScoreBasedGM}. Various guidance methods, including classifier-based \citep{Song2020ScoreBasedGM,Dhariwal2021DiffusionMB} and classifier-free techniques \citep{Ho2022ClassifierFreeDG}, have been developed. Recent efforts aim to refine control styles \citep{Bansal2023UniversalGF, Yu2023FreeDoMTE} and enhance sampling quality \citep{Hong2022ImprovingSQ, Kim2022RefiningGP}. However, current guidance methods suffer from quality degradation and color saturation issues at high guidance scales. Dynamic thresholding \cite{saharia2022photorealistic} attempts to address this by normalizing quantiles of pixels, but its effectiveness is limited to color-based image pixels, falling short in general tasks like latent space sampling. To suppress artifacts, community later had to use "scale mimic" that mixes dynamic thresholded latents with low guidance scale latents \cite{mcmonkeyprojects2024sddynamic}. Instead of these workarounds, our research aims to address quality issues of classifier-free guidance theoretically and systematically for general tasks, by tackling the non-linear aspects of DDPMs.

\subsection{Fast Sampling Methods} The significant challenge posed by the slow sampling speed of DDPMs, requiring numerous model evaluations, has spurred the development of accelerated sampling strategies. These strategies range from diffusion process approximations \citep{Song2020ScoreBasedGM, Liu2022PseudoNM, Song2020DenoisingDI, Zhao2023UniPCAU} to employing advanced solvers like Runge–Kutta and predictor-corrector methods \citep{Karras2022ElucidatingTD, Lu2022DPMSolverFS, Zhao2023UniPCAU}. These methods primarily utilize scores or predicted noises for de-noising. Our research focuses on designing a high guidance scale correction that seamlessly integrates with these fast sampling methods, primarily through the modification of scores or predicted noises.

\section{Background}
\label{Background}
\subsection{DDPM and the FP Equation}\label{Background DDPM}

DDPM models distributions of images $p(\mathbf{x})$ by recovering an original image $\mathbf{x}_0 \sim p(\mathbf{x})$ from one of its noise contaminated versions $\mathbf{x}_i$. The noise contaminated image is a linear combination of the original image and a gaussian noise
\begin{equation} \label{Background 1.5 forward diffusion process sampling discrete, alt ss}
   \mathbf{x}_{i} 
=\sqrt{\bar{\alpha}_i}\mathbf{x}_0 + \sqrt{1-\bar{\alpha}_i}\bar{\boldsymbol{\epsilon}}_i; \quad 1\le i \le n,
\end{equation}
where $n$ is the total diffusion steps, $\bar{\alpha}_i$ is the contamination weight at a time $t_i \in [0,T]$ in the forward diffusion process, and $\bar{\boldsymbol{\epsilon}}_i$ is a standard Gaussian random noise. More detailed information can be found in Appendix \ref{App1}.

The DDPM trains a denoising neural network $\boldsymbol{\epsilon}_\theta( \mathbf{x}, t_i  )$ to predict and remove the noise $\bar{\boldsymbol{\epsilon}}_i $ from $ \mathbf{x}_{i}$. It minimizes the denoising objective \citep{Ho2020DenoisingDP}:
\begin{equation} \label{Background 1.6 DDPM objective}
     L(\boldsymbol{\epsilon}_\theta) =\frac{1}{n}\sum_{i=1}^n \mathbf{E}_{\mathbf{x}_0 \sim p(\mathbf{x}), \bar{\boldsymbol{\epsilon}}_i\sim \mathcal{N}(\mathbf{0},I)}\| \bar{\boldsymbol{\epsilon}}_i  -  \boldsymbol{\epsilon}_\theta( \bold{x}_i, t_i  )\|_2^2.
\end{equation} This neural network is later used in the backward diffusion process to generate image samples from pure Gaussian noises. In the ideal scenario where diffusion time steps among $t_i$ are \textbf{infinitesimal}, the optimal solution for \eqref{Background 1.6 DDPM objective} is defined by:
\begin{equation} \label{optimal solution eps}
\boldsymbol{\epsilon}(\mathbf{x},t) = \argmin_{\boldsymbol{\epsilon}_\theta} L(\boldsymbol{\epsilon}_\theta) .
\end{equation} 
The forward process \eqref{Background 1.5 forward diffusion process sampling discrete, alt ss} places stringent constraints on its permissible forms. First, $\boldsymbol{\epsilon}( \mathbf{x}, t_i  )$ is proportional to the score function $\mathbf{s}(\mathbf{x}, t_i)=\nabla_{\mathbf{x}_{i}} \log p(\mathbf{x}_{i})$ \cite{Yang2022DiffusionMA} (Appendix \ref{App1}). Second, the score function is a solution of the score FP equation \cite{Lai2022FPDiffusionIS}, which could be rewritten (Appendix \ref{App3}) in terms of $\boldsymbol{\epsilon}(\mathbf{x},t)$ as
\begin{equation} \label{1.7 OU Fokker Plank of score}
  \frac{\partial \boldsymbol{\epsilon} }{\partial t}  =  \frac{1}{2} \left(  \mathcal{L}\boldsymbol{\epsilon}   -    \frac{ 1 }{ \sigma(t)}  \nabla_\mathbf{x} \| \boldsymbol{\epsilon}   \|^2_2\right),
\end{equation}
in which $\mathcal{L}\boldsymbol{\epsilon} = \nabla_\mathbf{x} (\boldsymbol{\epsilon}\cdot \mathbf{x})  +    \nabla_\mathbf{x}^2 \boldsymbol{\epsilon}  +  \frac{1-\sigma(t)^2}{\sigma(t)^2  } \boldsymbol{\epsilon} $ and $\sigma(t) = \sqrt{1-e^{-t}}$. These constraints lay the foundation for the duality between forward and backward diffusion processes, which is essential for successful sampling from $p(\mathbf{x})$.

\subsection{Conditional DDPM and Classifier-Free Guidance}

Conditional DDPMs, which generate images based on a given condition $\mathbf{c}$, model the conditional distribution $p(\mathbf{x} | \mathbf{c})$ with a denoising neural network represented as $\boldsymbol{\epsilon}_\theta(\mathbf{x}|\mathbf{c}, t_i)$. However, the training data in practice might only have weak or noised information about condition $\mathbf{c}$, therefore we need a way to enhance the control strength in this situation.

Guidance \cite{Song2020ScoreBasedGM,Dhariwal2021DiffusionMB,Ho2022ClassifierFreeDG} is a technique for conditional image generation that trades off control strength and image diversity. It generally aims to sample from the distribution:
\begin{equation} \label{Background guided diffusion p}
    p(\mathbf{x} | \mathbf{c}, \omega) \propto p(\mathbf{x} | \mathbf{c})^{1+\omega}p(\mathbf{x})^{-\omega} \propto p(\mathbf{c} | \mathbf{x})^{1+\omega}p(\mathbf{x}),
\end{equation}
where $\omega > 0$ is the guidance scale. When $\omega$ is large, this distribution concentrates on samples that have the highest conditional likelihood $p(\mathbf{c} | \mathbf{x})$. 


Classifier free guidance \citep{Ho2022ClassifierFreeDG} use the following guided denoising neural network $\boldsymbol{\epsilon}_{CF}$, deduced from \eqref{Background guided diffusion p} using $\boldsymbol{\epsilon} \propto \nabla \log p$, to approximately sample from $p(\mathbf{x} | \mathbf{c}, \omega)$: 
\begin{equation} \label{Background cfg eps}
    \boldsymbol{\epsilon}_{CF}(\mathbf{x}|\mathbf{c}, t_i , \omega) =(1+\omega) \  \boldsymbol{\epsilon}_\theta(\mathbf{x}|\mathbf{c}, t_i)-\omega \  \boldsymbol{\epsilon}_\theta(\mathbf{x}, t_i).
\end{equation}
It exactly computes $\boldsymbol{\epsilon}(\mathbf{x}|\mathbf{c}, t_i, \omega)$, the de-noising neural network of $p(\mathbf{x} | \mathbf{c}, \omega)$, at time $t_i=0$ (Appendix \ref{App2}). However, it is not a good approximation for $t_i > 0$ as it deviates from the FP equation \eqref{1.7 OU Fokker Plank of score} when $\omega$ is large. 

\subsection{Guidance's Deviation from the FP Equation}

We measure the deviation of a guidance method from the FP equation using the \textbf{mixing error}:
\begin{equation} \label{Background mixing error}
       \mathbf{e}_{m}(\boldsymbol{\epsilon},\mathbf{x},t) =  \frac{\partial \boldsymbol{\epsilon} }{\partial t}  -  \frac{1}{2} \left(  \mathcal{L}\boldsymbol{\epsilon}   -    \frac{ \nabla_\mathbf{x} \| \boldsymbol{\epsilon}   \|^2_2 }{ \sigma(t)}  \right).
\end{equation}
in which $\boldsymbol{\epsilon}$ is the guided denoising neural network under \textbf{infinitesimal} diffusion time steps.

For classifier-free guidance, the mixing error comes from the non-linear term of \eqref{1.7 OU Fokker Plank of score} as
\begin{equation} \label{Background mixing error CF}
\begin{split}
       &\mathbf{e}_{m}(\boldsymbol{\epsilon}_{CF},\mathbf{x},t) = \frac{1}{\sigma(t)} \nabla_\mathbf{x} (  \|  \boldsymbol{\epsilon}_{CF}(\mathbf{x}|\mathbf{c}, t_i , \omega) \|^2_2 - \\&  (1+\omega) \|   \boldsymbol{\epsilon}_\theta(\mathbf{x}|\mathbf{c}, t_i) \|^2_2 + \omega \|  \boldsymbol{\epsilon}_\theta(\mathbf{x}, t_i) \|^2_2 ),
\end{split}
\end{equation}
whose amplitude escalates with an increase in $\omega$ and further intensifies as $\sigma(t)$ decreases. 

\subsection{The Method of Characteristics} \label{MOC}

The method of characteristics \cite{evans2022partial} provides an analytical approach to solving certain kind of non-linear partial differential equations (PDE). It employs characteristic lines, a family of time-dependent trajectories defined by $\mathbf{x}_t = \mathbf{x}(t)$, along which the equation
\begin{equation}
    \boldsymbol{\epsilon}(\mathbf{x}_t,t) = \mathbf{F}_t( \boldsymbol{\epsilon}(\mathbf{x}_0,0) )
\end{equation}
holds, where $\boldsymbol{\epsilon}$ is a solution of the PDE and $\mathbf{F}_t$ is a deterministic and invertible function. This implies that the value of $\boldsymbol{\epsilon}(\mathbf{x}_t,t)$ can be inferred from its initial condition $\boldsymbol{\epsilon}(\mathbf{x}_0,0)$ at a specific point $\mathbf{x}_0$, determined by the characteristic line itself. 

The method of characteristics is applicable to \eqref{1.7 OU Fokker Plank of score} under certain assumptions, whose characteristic lines are detailed in \eqref{App5 charaline} of Appendix \ref{App5}. Particularly, the function $\mathbf{F}_t$ for \eqref{1.7 OU Fokker Plank of score} is a linear function.

\section{Problem Description}

Our method emphasizes the importance of adhering to the Fokker-Planck (FP) equation for two main reasons: Theoretically, the deviation from the FP equation leads to mixing errors, disrupting the equivalence between forward and backward diffusion processes and resulting in sample generation irregularities. Empirically, as we will shown in Sec.\ref{sec:gaussian_toy}, aligning with the FP equation helps to remove irregularities of classifier-free guidance.

We focus on a guided DDPM that generates samples from $p(\mathbf{x} | \mathbf{c}, \omega) $ \eqref{Background guided diffusion p}, constructing its denoising neural network $\boldsymbol{\epsilon}_{CH}(\mathbf{x}|\mathbf{c}, t_i , \omega)$ from two known networks: $\boldsymbol{\epsilon}_\theta(\mathbf{x}|\mathbf{c}, t_i)$ and $\boldsymbol{\epsilon}_\theta(\mathbf{x}, t_i)$.
The desired properties of $\boldsymbol{\epsilon}_{CH}$ include:
\begin{itemize}
    \item $\boldsymbol{\epsilon}_{CH}$ has no mixing error (satisfying FP equation);
    \item $\boldsymbol{\epsilon}_{CH}$ requires no training.
\end{itemize}

Constructing such $\boldsymbol{\epsilon}_{CH}$ is highly non-trivial as it requires tackling the non-linear term in the Fokker-Planck equation.

\section{Methodology}
\label{Methogology}
\subsection{The Characteristic Guidance} \label{ch guidance}
We propose the characteristic guidance as non-linear corrected classifier-free guidance:
\begin{equation} \label{Method characteristic guidance eps}
\boldsymbol{\epsilon}_{CH}(\mathbf{x}|\mathbf{c}, t_i , \omega) = (1+\omega) \  \boldsymbol{\epsilon}_\theta(\mathbf{x}_1|\mathbf{c}, t_i )-\omega \  \boldsymbol{\epsilon}_\theta(\mathbf{x}_2, t_i),
\end{equation}
in which $\mathbf{x}_1 = \mathbf{x} + \omega  \Delta \mathbf{x}$, $\mathbf{x}_2 = \mathbf{x}+ (1+\omega)\Delta \mathbf{x}$, and $\Delta \mathbf{x}$ is a non-linear correction term. It is evident that when $\Delta \mathbf{x}=0$, the characteristic guidance is equivalent to the classifier-free guidance \eqref{Background cfg eps}.

The correction $\Delta \mathbf{x}$ is determined from the training-free non-linear relation:
\begin{equation} \label{non-linear delta relation}
        \Delta \mathbf{x} =\mathbf{P} \circ \left( \boldsymbol{\epsilon}_\theta( \mathbf{x}_2, t_i  ) -\boldsymbol{\epsilon}_\theta(\mathbf{x}_1 |\mathbf{c}, t_i  ) \right) \sigma_i,
\end{equation}
where $\sigma_i = \sqrt{ 1-\bar{\alpha}_i }$ is a scale parameter and the operator $\mathbf{P}$ could be an arbitrary orthogonal projection operator. 

Equation \eqref{non-linear delta relation} can be solved using the fixed-point iteration method \cite{evans2022partial} (Appendix \ref{App4}). For practical efficiency, we propose accelerated fixed-point iteration algorithms, Alg.\ref{RMS ite} and Alg.\ref{AA ite}, to minimize the iterations needed for convergence.

The projection operator $\mathbf{P}$ should be theoretically the identity, but we found that orthogonal projections that serve as regularization greatly accelerates the computation. In practice, $\mathbf{P}$ acts channel-wisely and is specified by a vector $\mathbf{g}$: 
\begin{equation} \label{projection P}
        \mathbf{P}_{\mathbf{g}} \circ \mathbf{v} = \frac{\mathbf{g} \cdot \mathbf{v}}{\mathbf{g} \cdot \mathbf{g}} \mathbf{g}
\end{equation}
For pixel space diffusion model, we suggest the operator $\mathbf{g} = \mathbf{1}$ as projection to the channel-wise mean. For latent space diffusion model, we suggest $\mathbf{g} = \left( \boldsymbol{\epsilon}_\theta( \mathbf{x}, t_i  ) -\boldsymbol{\epsilon}_\theta(\mathbf{x} |\mathbf{c}, t_i  ) \right) \sigma_i$. For low dimensional cases that are not images, we suggest the operator $\mathbf{P}$ to be identity.

 \subsection{Theoretical Foundation}

This section derives the characteristic guidance \eqref{Method characteristic guidance eps}-\eqref{non-linear delta relation} by solving the FP equation \eqref{1.7 OU Fokker Plank of score} under the Harmonic ansatz. We propose the Harmonic ansatz assuming that the optimal DDPM solution $\epsilon(\mathbf{x},t)$ of \eqref{optimal solution eps} is harmonic:
\begin{ansatz}\label{ansatz H}
The Harmonic Ansatz: The optimal solution $\boldsymbol{\epsilon}(\mathbf{x},t)$ is harmoic
\begin{equation}  \nabla_{\mathbf{x}}^2  \boldsymbol{\epsilon}(\mathbf{x},t) = 0.\end{equation} 
\end{ansatz}
This ansatz is inspired by and holds exactly when the distribution of original images 
 $p(\mathbf{x})$ is a Gaussian distribution, and we will show that it works for other cases as well in experiments \ref{sec:gaussian_toy} and \ref{cool mag}. 
 
Characteristic guidance adopts the same formulation as classifier-free guidance at $t=0$
\begin{equation} \label{CHG eps 0}
    \boldsymbol{\epsilon}_{CH}(\mathbf{x}_0|\mathbf{c}, 0 , \omega) =(1+\omega) \  \boldsymbol{\epsilon}(\mathbf{x}_0|\mathbf{c},  0)-\omega \  \boldsymbol{\epsilon}(\mathbf{x}_0,  0),
\end{equation}
in which parameters $\theta$ are omitted since we are considering optimal solutions with infinitesimal time steps. This guarantees that $\boldsymbol{\epsilon}_{CH}$ samples from $p(\mathbf{x} | \mathbf{c}, \omega)$, which aligns with the target of classifier-free guidance. 

Next, our aim is to construct the unknown $\boldsymbol{\epsilon}_{CH}$ for any time $t>0$ using the known $\boldsymbol{\epsilon}(\mathbf{x}|\mathbf{c}, t)$ and $\boldsymbol{\epsilon}(\mathbf{x}, t)$, by solving the Fokker-Planck equation \eqref{1.7 OU Fokker Plank of score} analytically. 

The harmonic ansatz \ref{ansatz H} simplifies the Fokker-Planck equation \eqref{1.7 OU Fokker Plank of score} into a quasi-linear equation, which can be solved by the method of characteristics in Sec.\ref{MOC}. Assuming $\boldsymbol{\epsilon}_{CH} (\mathbf{x}|\mathbf{c}, t, \omega)$, $\boldsymbol{\epsilon} (\mathbf{x}|\mathbf{c}, t)$, and $\boldsymbol{\epsilon} (\mathbf{x}, t)$ are three solutions of \eqref{1.7 OU Fokker Plank of score}, there exist three distinct characteristic lines $\mathbf{x}(t)$, $\mathbf{x}^{(1)}(t)$, and $\mathbf{x}^{(2)}(t)$ passing through the point $\mathbf{x}_0$ at $t=0$, satisfying
\begin{equation}
    \begin{split}
   \boldsymbol{\epsilon}_{CH}(\mathbf{x}_0|\mathbf{c}, 0, \omega) &= \mathbf{F}_t^{-1}\left[\boldsymbol{\epsilon}_{CH}(\mathbf{x}_t |\mathbf{c}, t, \omega)  \right], \\
 \boldsymbol{\epsilon}(\mathbf{x}_0|\mathbf{c}, 0) &= \mathbf{F}_t^{-1}\left[ \boldsymbol{\epsilon}(\mathbf{x}_t^{(1)}|\mathbf{c}, t) \right], \\
 \boldsymbol{\epsilon}(\mathbf{x}_0, 0) &= \mathbf{F}_t^{-1}\left[ \boldsymbol{\epsilon}(\mathbf{x}_t^{(2)}, t) \right],
    \end{split}
\end{equation}
in which the function $\mathbf{F}_t$ for characteristic lines of \eqref{1.7 OU Fokker Plank of score} is a linear function. Consequently, these characteristic lines and the initial condition \eqref{CHG eps 0} yield an elegant analytical solution of $\boldsymbol{\epsilon}_{CH}$:
\begin{equation}
\boldsymbol{\epsilon}_{CH} (\mathbf{x}_t|\mathbf{c}, t , \omega) = (1+\omega) \ \boldsymbol{\epsilon}(\mathbf{x}_t^{(1)}|\mathbf{c}, t) - \omega \ \boldsymbol{\epsilon}(\mathbf{x}_t^{(2)}, t).
\end{equation}
This result establishes the possibility of constructing $\boldsymbol{\epsilon}_{CH}$ using $\boldsymbol{\epsilon}_\theta(\mathbf{x}|\mathbf{c}, t_i)$ and $\boldsymbol{\epsilon}_\theta(\mathbf{x}, t_i)$. Its precise form is described in the following lemma: 
\begin{lemma} \label{lemma chara}
Let $\boldsymbol{\epsilon}(\mathbf{x}, t)$, $\boldsymbol{\epsilon}_1(\mathbf{x}, t)$, and $\boldsymbol{\epsilon}_2(\mathbf{x}, t)$ be three distinct solutions of the FP equation \eqref{1.7 OU Fokker Plank of score}, satisfying the Harmonic Ansatz \ref{ansatz H}. Moreover, their initial condition satisfies
\begin{equation} \label{lemma ini}
\boldsymbol{\epsilon}(\mathbf{x}, 0) = (1+\omega)\boldsymbol{\epsilon}_1(\mathbf{x}, 0) -\omega \boldsymbol{\epsilon}_2(\mathbf{x}, 0),
\end{equation}
Then, we have the relation
\begin{equation} 
\boldsymbol{\epsilon}(\mathbf{x}, t) = (1+\omega)\boldsymbol{\epsilon}_1(\mathbf{x} + \omega \Delta \mathbf{x}, t) -\omega \boldsymbol{\epsilon}_2(\mathbf{x}+(1+\omega)\Delta \mathbf{x}, t),
\end{equation}
where $\Delta \mathbf{x}$ is given by
\begin{equation} 
\Delta \mathbf{x} = \left( \boldsymbol{\epsilon}_2( \mathbf{x}+(1+\omega)\Delta \mathbf{x}, t  ) -\boldsymbol{\epsilon}_1( \mathbf{x} + \omega \Delta \mathbf{x}, t  ) \right) \sigma(t),
\end{equation}
in which $\sigma(t) = \sqrt{1-e^{-t}}$.
\end{lemma}
The proof of lemma \ref{lemma chara} is established in Appendix \ref{App5}. Replacing the continuous functions $\sigma(t)$, $\boldsymbol{\epsilon}$, $\boldsymbol{\epsilon}_1$, and $\boldsymbol{\epsilon}_2$ with their discretized version $\sigma_i$, $\boldsymbol{\epsilon}_{CH}(\mathbf{x}|\mathbf{c}, t_i , \omega)$, $\boldsymbol{\epsilon}_\theta(\mathbf{x}|\mathbf{c}, t_i )$, and $\boldsymbol{\epsilon}_\theta(\mathbf{x}, t_i)$ gives us the characteristic guidance. Besides, a direct consequence of lemma \ref{lemma chara} is the following Theorem
\begin{theorem} \label{th chara}
The characteristic guidance has \textbf{no mixing error} when the Harmonic ansatz applies and diffusion time steps are infinitesimal:
\begin{equation} \label{Background mixing error ch}
       \mathbf{e}_{m}(\boldsymbol{\epsilon}_{CH},\mathbf{x},t) = 0.
\end{equation}
\end{theorem}

In summary, $\boldsymbol{\epsilon}_{CH}$ is an exact solution to the FP equation \eqref{1.7 OU Fokker Plank of score} with initial condition \eqref{CHG eps 0} under the harmonic ansatz, eliminating the mixing error caused by non-linearity of \eqref{1.7 OU Fokker Plank of score}.

\section{Experiments}
\subsection{Gaussians: Where the Harmonic Ansatz Holds}\label{sec:gaussian_toy}
\begin{figure}[ht]
\begin{center}
\centerline{\includegraphics[width=0.8\columnwidth]{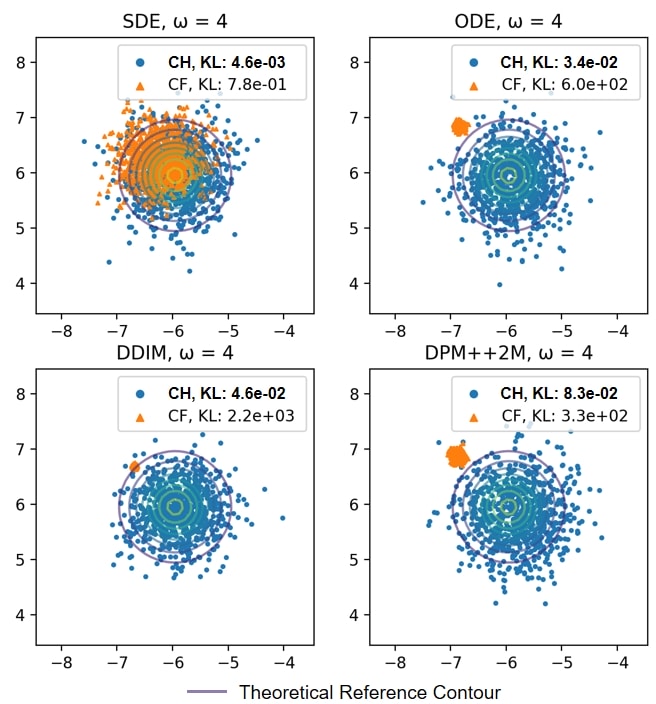}}
\caption{Samples and the KL divergence from characteristic guidance (CH) and classifier free guidance (CF) guided DDPM modeling conditional Gaussian distribution. The contours corresponds to the theoretical reference (ground truth) distribution of the guided DDPM \eqref{Background guided diffusion p}.} 
\label{conditional gaussian omega 4}
\end{center}
\vspace{-0.5cm}
\end{figure}

\begin{figure}[ht]
\begin{center}
\centerline{\includegraphics[width=0.8\columnwidth]{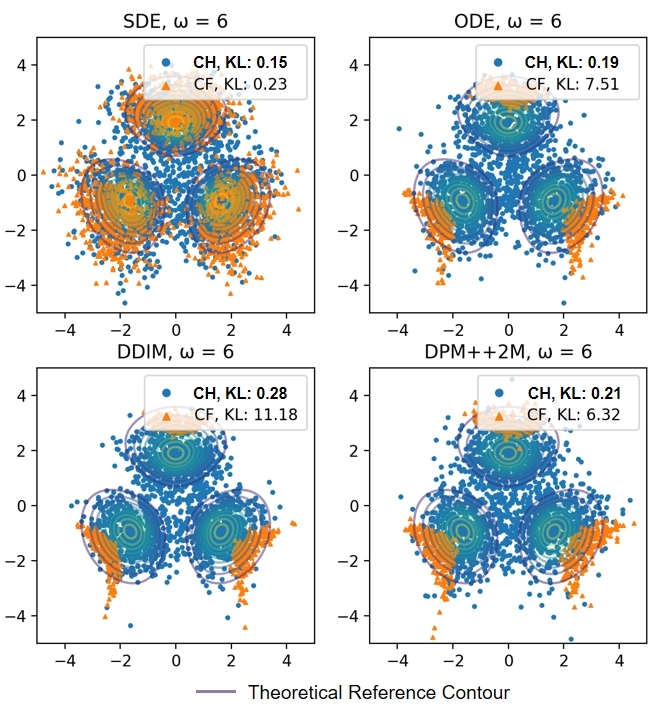}}
\caption{Comparison between CH (ours) and CF guided DDPM on modeling mixture of Gaussian distribution. The contours corresponds to the theoretical reference distribution of the guided DDPM \eqref{Background guided diffusion p}. Samples from characteristic guidance shows better KL divergence than those from classifier-free guidance.} 
\label{mixture gaussian omega 6}
\end{center}
\vspace{-1cm}
\end{figure}
This section demonstrates the theoretical advantage of characteristic guidance on Gaussian models, where we can compare samples against analytical solutions. We conducted experiments to compare classifier-free guidance and characteristic guidance under two scenarios: sampling from (1) conditional 2D Gaussian distributions, where the harmonic ansatz is exact; (2) mixture of Gaussian distributions, where the harmonic ansatz holds for most of the region except where the components overlap significantly. The guided diffusion model \eqref{guided diffusion p} aims to sample from the distribution \eqref{Background guided diffusion p} whose contour is plotted in Fig.\ref{conditional gaussian omega 4} and Fig.\ref{mixture gaussian omega 6}. 

In the conditional Gaussian scenario, we analytically set two DDPMs for the conditional distribution $p(\mathbf{x}|\mathbf{c})$ and the unconditional distribution $p(\mathbf{x})$ as follows:
\begin{equation} 
\begin{split}
p(\mathbf{x}|\mathbf{c})&=\mathcal{N}\left( x_1,x_2 |(c_1,c_2)^T,  I \right)\\  
p(\mathbf{x})&=\mathcal{N}\left( x_1,x_2 |(0,0)^T, 5 I \right). 
\end{split}
\end{equation}
where $c_1 = -5, c_2 = 5$ in the experiment.
In the mixture of Gaussian, we trained conditional and unconditional DDPMs to learn the distributions $p(\mathbf{x}|\mathbf{c})=  \prod_{i=0}^2 \mathcal{N}\left(\mathbf{x}|\boldsymbol{\mu}_i,  I \right)^{c_i}$ and $p(\mathbf{x})=\frac{1}{3} \sum_{\{\mathbf{e}_i,i=0,1,2\}} p(\mathbf{x}|\mathbf{e}_i)$,
where $\mathbf{c} = (c_0,c_1, c_2)^T$ represents a three dimensional one-hot vector, $\boldsymbol{\mu}_0 = (-1,-1/\sqrt{3})^T$, $\boldsymbol{\mu}_1 = (1,-1/\sqrt{3})^T$, $\boldsymbol{\mu}_2 = (0,\sqrt{3} -1/\sqrt{3})^T$, $\{\mathbf{e}_i,i=0,1,2\}$ represents all one-hot vectors in three dimension space.

In both experiments, we evaluated four different sampling methods:  SDE \eqref{1.5 backward diffusion process sampling discrete, alt} (1000 steps), probabilistic ODE \cite{Song2020ScoreBasedGM} (1000 steps), DDIM \cite{Song2020DenoisingDI} (20 steps), and DPM++2M \cite{Lu2022DPMSolverFS} (20 steps). The projection operator $\mathbf{P}$ for characteristic guidance was set to identity. Samples were compared against the theoretical reference using Kullback–Leibler (KL) divergence.

In the conditional Gaussian scenario, Fig.\ref{conditional gaussian omega 4} shows that characteristic guidance achieves better KL divergence than classifier-free guidance for every sampling method. It's worth noting that classifier-free guidance suffers from severe bias and catastrophic loss of diversity when ODE based sampling methods (ODE, DDIM, and DPM++2M) are used. Contrarily, characteristic guidance always yields correct sampling with KL divergence better than classifier-free guidance's best SDE samples. In the Mixture of Gaussian scenario, Fig.\ref{mixture gaussian omega 6} shows that characteristic guidance again showed less bias and better diversity. However, the overlap regions in the mixture of Gaussian scenario led to less clear component boundaries. This is because the harmonic ansatz is not valid in the overlap regions. Besides, minor artifacts appear because iterating on the trained neural network $\boldsymbol{\epsilon}$ brings in approximation errors. Overall, characteristic guidance outperformed classifier-free guidance in reducing bias, increasing diversity, and narrowing the performance gap between ODE and SDE sampling methods. This was consistently observed in cases where the harmonic ansatz holds exactly or approximately. More results could be found in Fig.\ref{The sampling results Gaussian} and Fig.\ref{The sampling results M Gaussian}.

\subsection{Cooling of the Magnet: Where the Harmonic Ansatz Not Supposed to Hold} \label{cool mag}
\begin{figure}[ht]
\vskip 0.in
\begin{center}
\centerline{\includegraphics[width=0.8\columnwidth]{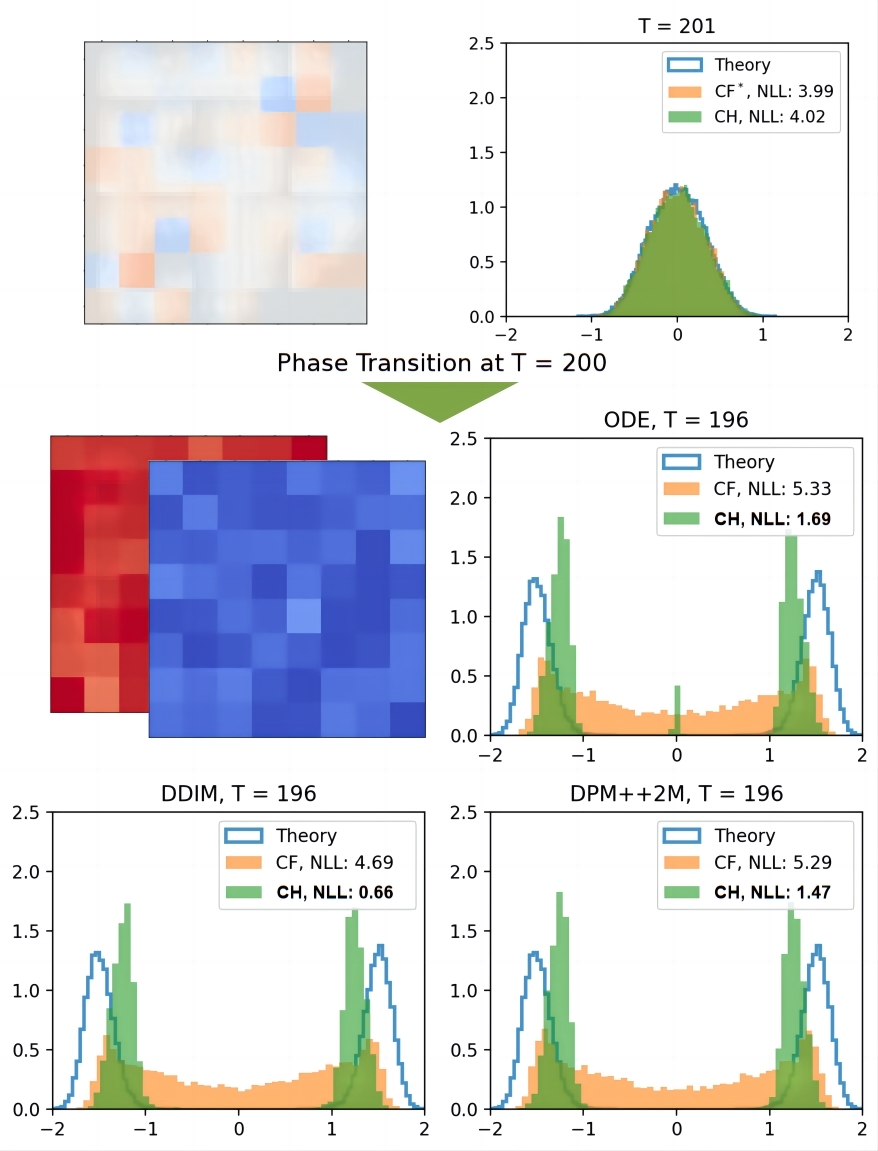}}
\caption{Comparison between CH (ours) and CF guided DDPM on simulating magnet cooling. The images in the upper and middle left (blue and red lattices) depict samples from DDPMs, while the histograms illustrate the distribution of samples' mean value (magnetization) across different temperatures. The contours corresponds to the theoretical reference distribution of magnetization. The characteristic guidance has better NLL and is more capable in capturing peak separation of sample magnetization.} 
\label{The sampling results phi4 196}
\end{center}
\vspace{-0.5cm}
\end{figure}

\begin{figure*}[ht]
\begin{center}
\includegraphics[width=2.1\columnwidth]{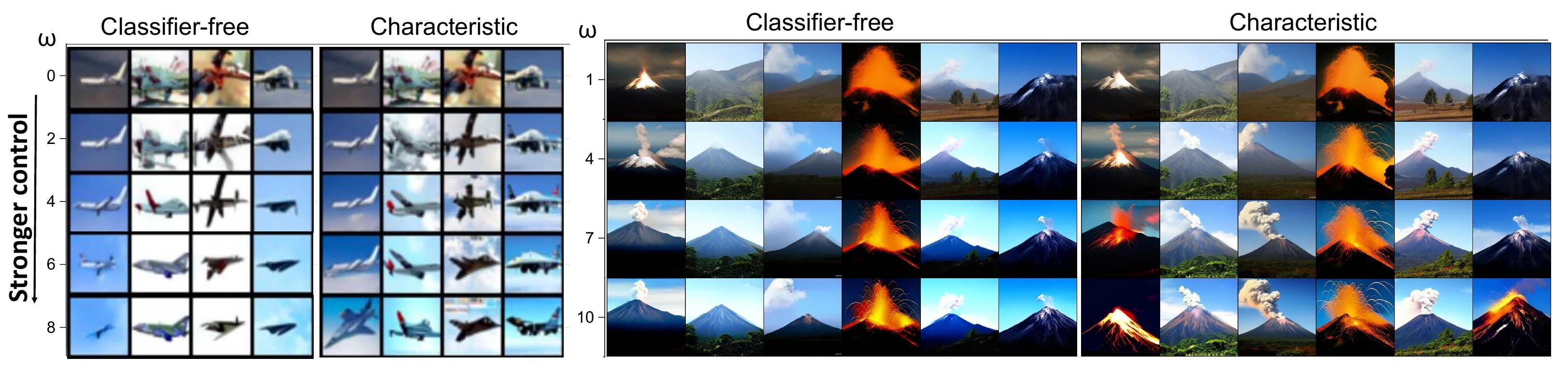}
\end{center}
\vspace{-0.5cm}
\caption{\textit{Left}: CIFAR-10 aircraft images generated via DDPM highlight the difference between Classifier-free Guidance (CF) and Characteristic Guidance (CH) across various guidance scales ($\omega$). The CF-guided images tend to have dull or even white backgrounds at higher $\omega$, whereas the CH-guided images creates more vibrant scenes with skies and clouds. \textit{Right}: In ImageNet 256, volcano samples generated using latent diffusion models with CF and CH Guidance without cherry-picking. CF images show color cast and underexposure at higher $\omega$, while CH images maintain consistent color and exposure, better highlighting volcanic features, such as smoke and lava. For these visual comparisons, consistent initial noise ensures that both CF and CH guided images maintain similar contexts at lower $\omega$ values.
}\label{Volcano}
\end{figure*}

In this section, we explore a simulation of the cooling of a magnet, comparing our method's performance with that of classifier-free guidance. This simulation is not just a representation of a complex real-world physical phenomenon but also a test of our characteristic guidance on a case where the harmonic ansatz does not hold.

We focus on the cooling around the Curie temperature, a critical point where a paramagnetic material undergoes a phase transition to permanent magnetism. This transition, driven by 3rd order terms in the score function, challenges the harmonic ansatz. Despite this, our method successfully captures the most important phenomenon of this transition: the separation of two distinct peaks below the Curie temperature, which classifier-free guidance fails to reproduce. These results emphasize the robustness and effectiveness of our method in modeling intricate physical processes, even in scenarios where the harmonic ansatz does not hold.


Suppose we have a square slice of 2-dimensional paramagnetic material which can be magnetized only in the direction perpendicular to it. It can be further divided uniformly into $8\times8$ smaller pieces, with each pieces identified by its index $(i,j)$. We use a scalar field $\phi_{i,j}$ to record the magnitude of magnetization of each piece, which can be viewed as a single channel $8\times8$ image of magnetization.

The probability of observing a particular configuration of $\phi$, according to statistical physics, is proportional to the Boltzmann distribution
\begin{equation} \label{Boltzmann distribution}
    p(\phi;T) \propto e^{-\beta H(\phi;T)},
\end{equation}
where $\beta H(\phi;T)$ is Hamiltonian at temperature $T$, assigning an energy to each of possible $\phi$. At temperature near the Curie temperature $T_c$, a Landau-Ginzburg model \cite{kardar_2007} of magnet uses a Hamiltonian \eqref{LG hamilton} (Appendix \ref{App8}) with 4-th order terms. Such terms are the key to characterise phase transition of magnet, but its gradient (score function) is a third order term that does not respect the harmonic ansatz.

A remarkable property about the Landau-Ginzburg model, bearing surprising similarity with the DDPM guidance \eqref{Background guided diffusion p}, is
\begin{equation}
    p(\phi;(1+\omega) T_1 - \omega T_0) \propto  p(\phi; T_1)^{1+\omega} p(\phi; T_0)^{-\omega},
\end{equation}
where $\omega$ is the guidance scale, $T_1 > T_0$ are two distinct temperature values. This means if we know the distribution of our magnet at two distinct temperatures $T_0$ and $T_1$, we know its distribution at any temperature cooler than $T_1$ by adjusting the guidance scale $\omega$.

We train a conditional DDPM simulating the cooling of a magnet with the Curie temperature $T_c=200$. The model is trained at two distinct temperatures $p(\phi|T_0=201)$ and $p(\phi|T_1=200)$, corresponds to $\omega = -1$ and $\omega = 0$. Samples at lower temperatures $T=196$ ($\omega = 4$) are later obtained by adjusting the guidance scale. Four distinct sampling methods, identical to those discussed in Sec.\ref{sec:gaussian_toy}, are evaluated. The projection operator $\mathbf{P}$ for characteristic guidance is set to be the channal-wise mean. 

Fig.\ref{The sampling results phi4 196} demonstrates the phase transition of mean magnetization, which presents the histograms of the mean value of $\phi$ sampled from the trained DDPM (8192 samples). Above the Curie temperature $T_c=200$, the histogram of the mean magnetization has one peak centered at 0. Below the Curie temperature, the histogram of the mean magnetization has two peaks with non-zero centers. More details are available in Fig.\ref{The sampling results phi4}. The change in the number of peaks represents a phase change from non-magnet to a permanent magnet. 

Characteristic guidance outperforms classifier-free guidance in sample quality below the Curie temperature, showcasing better negative log-likelihood (NLL) and effectively capturing distinct peaks in magnetization of samples. In contrast, classifier-free guidance struggles to produce distinct peaks. Despite its slightly biased peak positions compared with theoretical expectations, characteristic guidance still demonstrates superior performance in describing distinct peaks and modeling phase transitions.

\subsection{CIFAR-10: Natural Image Generation}
In this study, we evaluate the effectiveness of classifier-free (CF) and characteristic (CH) guidance in generating natural images using a DDPM trained on the CIFAR-10 dataset. We assess the quality of the generated images using Frechet Inception Distance (FID), Inception Score (IS), and visual inspection (Sec \ref{Visual Inspection}).

\begin{table}[h]
\caption{Comparison of FID and IS for CF and CH guided DDPM on the CIFAR-10 dataset, using DDIM (50 steps) and DPM++2M (50 steps). The operator $\mathbf{P}$ for CH is the projection from the input to its channel-wise mean. Characteristic guidance generally achieves lower FID and comparable IS, particularly at higher guidance scales $\omega$.}
\label{cifar-table}
\vskip 0.15in
\scalebox{0.8}{
\begin{tabular}{l|llll|llll}
\hline
\multirow{3}{*}{$\omega$} & \multicolumn{4}{c|}{DDIM}                         & \multicolumn{4}{c}{DPM++2M}                      \\ \cline{2-9} 
                   & \multicolumn{2}{c}{FID $\downarrow$} & \multicolumn{2}{c|}{IS $\uparrow$} & \multicolumn{2}{c}{FID $\downarrow$ } & \multicolumn{2}{c}{IS $\uparrow$} \\ \cline{2-9} 
                   & CF         &\multicolumn{1}{l|}{CH}            & CF        & CH            & CF             & \multicolumn{1}{l|}{CH}     & CF         & CH        \\ \hline
0.3                & 4.52       &\multicolumn{1}{l|}{\textbf{4.46}} & 9.18      & \textbf{9.23} & \textbf{3.33}  &\multicolumn{1}{l|}{3.35}    & 9.69       &  9.69         \\
0.6                & 4.80       &\multicolumn{1}{l|}{\textbf{4.64}} & 9.39      & \textbf{9.43} &  3.51          &\multicolumn{1}{l|}{\textbf{3.44}}    & 9.93       &  9.93        \\
1.0                & 6.22       &\multicolumn{1}{l|}{\textbf{5.86}} & 9.52      & \textbf{9.53} &  4.75          &\multicolumn{1}{l|}{\textbf{4.51}}    & \textbf{10.06}      &  10.04         \\
1.5                & 8.56       &\multicolumn{1}{l|}{\textbf{7.89}} & 9.50      & \textbf{9.57} &  6.85          &\multicolumn{1}{l|}{\textbf{6.37}}    & \textbf{10.08}      &  10.07         \\ 
2.0                & 11.02      &\multicolumn{1}{l|}{\textbf{10.10}}& 9.49      & \textbf{9.51} &  9.11          &\multicolumn{1}{l|}{\textbf{8.34}}    & 10.04      &  10.04        \\ 
4.0                & 19.85      &\multicolumn{1}{l|}{\textbf{18.15}}& \textbf{9.23}     & 9.22  &  17.04         &\multicolumn{1}{l|}{\textbf{15.52}}   & 9.75       &  \textbf{9.76}       \\ 
6.0                & 27.04      &\multicolumn{1}{l|}{\textbf{24.77}}& 8.85      & \textbf{8.86} &  23.47         &\multicolumn{1}{l|}{\textbf{21.46}}   & 9.33       &  \textbf{9.38}        \\ 
\hline
\end{tabular}}
\vspace{-0.1in}
\end{table}
Our experiments, detailed in Table.\ref{cifar-table}, demonstrate that characteristic guidance achieves a lower FID while maintaining a comparable IS to classifier-free guidance, illustrating its effectiveness in balancing control strength and diversity. Unlike the typical trade-off between FID and IS in guidance methods \cite{Ho2022ClassifierFreeDG}, where enhancing diversity marked by improved FID often reduces control strength by a decrease in IS, characteristic guidance improves FID without compromising control strength. These results underscore its advantage in producing high-fidelity images while preserving both diversity and control.

\subsection{ImageNet 256: Correction for Latent Space Model}

The characteristic guidance's effect on latent space diffusion models (LDM) \citep{Rombach2021HighResolutionIS} is different: it enhances the control strength rather than lowering the FID. We test the characteristic guidance on the LDM for ImageNet256 dataset, with codes and models adopted from the LDM paper \citep{Rombach2021HighResolutionIS}. The quality of generated images are evaluated by FID, IS and visual inspection (Sec \ref{Visual Inspection}). 

\begin{figure}[ht]
\begin{center}
\centerline{\includegraphics[width=0.9\columnwidth]{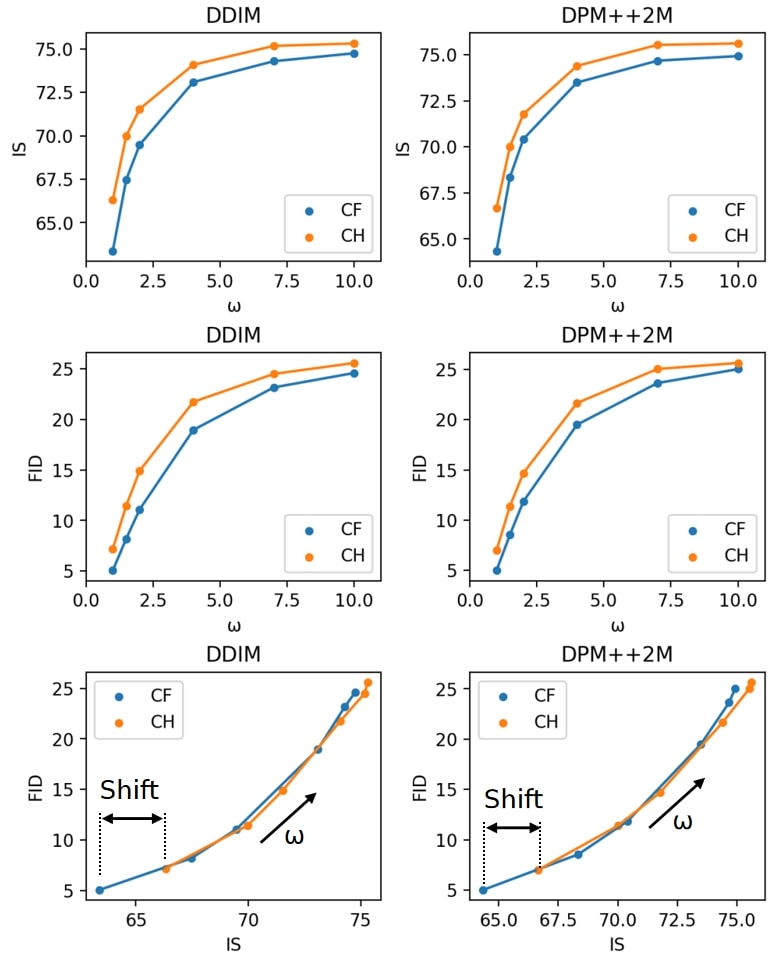}}
\caption{FID and IS comparison between CH and CF guided DDPM on ImageNet 256, using DDIM and DPM++2M sampling methods. CH shows better IS but higher FID than CF, with a rightward shift in its IS-FID curve indicating higher effective guidance scale. } 
\label{ldm plot}
\end{center}
\vspace{-1cm}
\end{figure}

Fig.\ref{ldm plot} presents sample quality metrics using DDIM (50 steps) and DPM++2M (50 steps). While CH improves IS, it adversely affects FID compared to CF. This IS-FID trade-off complicates performance evaluation. Notably, the IS-FID curves of CH and CF align closely, with CH demonstrating a rightward shift. This means CH guidance attains a certain IS with lower guidance scale, suggesting enhanced control strength.

The difference in CH guidance's functionality between CIFAR-10 and ImageNet256 may relate to the choice of projection operator $\mathbf{P}$. In latent space, we apply channel-wise projection to the residual vector, avoiding channel-wise mean projection inappropriate for RGB color correction in latent space. Despite the unclear interpretation of this approach in latent space, our visual inspection suggest CH guidance achieves semantic characteristics enhancing and exposure correction, as shown in Fig.\ref{Volcano} (Right).


\subsection{Stable Diffusion: Text/Image to Image Generation}
Characteristic guidance, designed for broad compatibility, can be seamlessly integrated with any DDPM that supports classifier-free guidance. We have successfully integrated our characteristic guidance into Stable Diffusion WebUI \cite{stable_diffusion_webui} as public available extension, supporting all provided samplers under both Txt2Img and Img2Img mode. Fig.\ref{SdXL} showcases a comparative visualization of images from Stable Diffusion XL \cite{podell2023sdxl}. More visulizations could be found in Appendix. 

\subsection{Visual Inspection} \label{Visual Inspection}

This section evaluates images generated using classifier-free (CF) and characteristic (CH) guidance from the same initial noise, demonstrating that CH guidance is more capable in enhancing semantic characteristics.

For aircraft images created by the Cifar-10 model, as shown in Fig.\ref{Volcano} (Left), CF guidance at higher scales ($\omega$) produces backgrounds that are often dull or white. In contrast, CH guidance results in more vivid and realistic backgrounds, such as skies with clouds that open appears in aircrafts photos. 
Similarly, with volcano images from ImageNet 256's latent diffusion models (Fig.\ref{Volcano}, Right), CF-guided images at elevated $\omega$ levels exhibit color casting and underexposure. CH guidance, conversely, maintains consistent color and exposure, better accentuating key volcanic elements like smoke and lava, thus emphasizing the semantic characteristics more effectively.

In stable diffusion scenarios, Fig.\ref{SdXL} compares CF and CH guidance across various prompts and CFG scales. At a CFG scale of 10 ($\omega = 9$), CH guidance mitigates anatomical irregularities, as seen in the handstand girl images (Fig.\ref{SdXL} Left). Increasing the scale to 30 ($\omega = 29$), CF guidance struggles with color cast and exposure issues, while CH guidance produces images that more aptly emphasize prompt-relevant features, such as running girls and lush mountains (Fig.\ref{SdXL} Right).

Overall, these comparisons underscore CH guidance's capability in not just adjusting basic elements like color and exposure, but more importantly, in amplifying the semantic characteristics of the subjects in line with the given prompts.

\section{Discussion}
\subsection{The Stopping Criteria for Iteration.}
Characteristic guidance applies non-linear correction at each time step by iteratively solving \eqref{non-linear delta relation}. The iteration process is governed by two key stopping criteria: a predefined threshold (tolerance $\eta$, detailed in Appendix.\ref{App4}) and a maximum iteration limit set at 10. We investigate the impact of the threshold on evaluation metrics (FID and IS) and the total number of iterations, which is crucial for computational efficiency as each iteration involves calling the de-noising neural network.

\begin{figure}[ht]
\vskip 0.in
\begin{center}
\centerline{\includegraphics[width=\columnwidth]{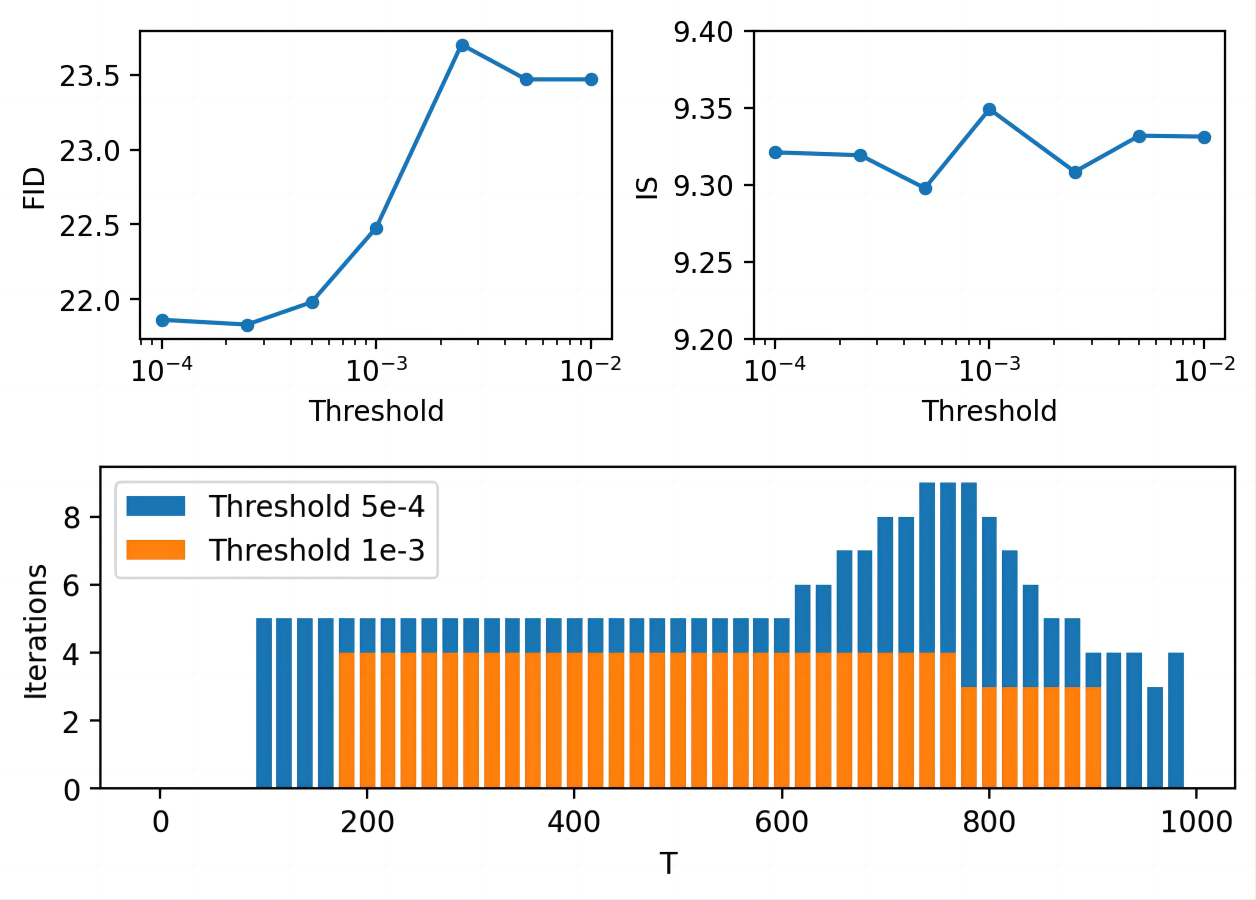}}
\caption{Analysis of convergence and iterations in characteristic guidance on the CIFAR-10 dataset ($\omega=6$, DPM++2M, 50 steps). The graph depicts the influence of varying threshold levels on FID and IS, and the iteration count across different thresholds relative to the diffusion time step $T$, ranging from $0$ (image) to $1000$ (noise).} 
\label{ite plot}
\end{center}
\vspace{-1.cm}
\end{figure}

As illustrated in Fig.\ref{ite plot}, we observe that convergence typically occurs at thresholds below $1e^{-3}$. This finding informs the optimal setting of $\eta$ to balance quality and efficiency. Additionally, the graph also highlights the iteration frequency employed by characteristic guidance across two threshold levels. It is noteworthy that non-linear corrections predominantly occur in the middle stages of the diffusion process, a phase critical for context generation as identified in \cite{Zhang2023ShiftDDPMsEC}.

\subsection{The Difference with Dynamical Thresholding}

Dynamical thresholding \cite{saharia2022photorealistic} is a technique designed for RGB images. It predicts the final output image $\mathbf{x}_0$ with \eqref{Background 1.5 forward diffusion process sampling discrete, alt ss} at each step, correcting and normalizing any out-of-range pixels back to $[-1,1]$. 

Dynamical threshold is effective in correct color cast when pixel values representing RGB colors. But it falls short in general tasks like latent space sampling and correcting contextual inaccuracies.
In contrast, characteristic guidance is a theoretical method that is applicable to any continuous data type of any dimension. In practice, as Fig.\ref{gymhandstand} shows, characteristic guidance is capable of correcting contextual inaccuracies while dynamical thresholding focuses more on color.

\section{Conclusion}
We introduced characteristic guidance, a novel guidance method providing non-linear correction to classifier-free guided DDPMs. It is, as far as we know, the first attempt to lay the theoretical framework for large guidance scale correction. Our comprehensive evaluations of characteristic guidance covered various scenarios, each with a distinct objective: validating its theoretical advantage on Gaussian models, applying it to physics problems like the Landau-Ginzburg model for magnetism, and demonstrating its robustness in image generation across different spaces — CIFAR-10 in pixel-space, ImageNet and Stable Diffusion in latent space. The results consistently highlight the method's effectiveness in enhancing the semantic characteristics of subjects and mitigating generation irregularities.

\section{Limitation} 

\noindent\textbf{FID's Effectiveness.}  The effectiveness of FID for evaluating guided DDPM is compromised by the difference between the target conditional probabilities \( p(\mathbf{x} | \mathbf{c}, \omega) \) and marginal probabilities \( p(\mathbf{x}) \). FID's focus on the distance from \( p(\mathbf{x}) \) contrasts with guided DDPM's goal of sampling from \( p(\mathbf{x} | \mathbf{c}, \omega) \), suggesting the need for more apt metrics that address this discrepancy.

\noindent\textbf{Speed.} The process of resolving the non-linear correction \eqref{non-linear delta relation} requires iterative computations involving neural networks. This approach, without effective regularization, tends to be slow. Exploring advanced regularization techniques beyond the projection $\mathbf{P}$ could enhance the convergence rate hence accelerate the computation.

\section*{Software and Data}

Characteristic guidance is available at \url{https://scraed.github.io/CharacteristicGuidance/} as open source and public extension for Stable Diffusion WebUI \cite{stable_diffusion_webui}, supporting all provided samplers under both Txt2Img and Img2Img mode.

\section*{Acknowledgements}

We would like to express our gratitude to Professor Yang Wang at the Hong Kong University of Science and Technology for his support of this study.

\section*{Impact Statement}

This paper presents work whose goal is to advance the field of 
Machine Learning. There are many potential societal consequences 
of our work, none which we feel must be specifically highlighted here.

\nocite{langley00}

\bibliography{example_paper}
\bibliographystyle{icml2024}

\newpage
\appendix
\onecolumn

\section{The Denoising Diffusion Probabilistic Model (DDPMs)} \label{App1}
DDPMs \citep{Ho2020DenoisingDP} are models that generate high-quality images from noise via a sequence of denoising steps. Denoting images as random variable $\mathbf{x}$ of the probabilistic density distribution $p(\mathbf{x})$, the DDPM aims to learn a model distribution that mimics the image distribution $p(\mathbf{x})$ and draw samples from it. The training and sampling of the DDPM utilize two diffusion process: the forward and the backward diffusion process. 

The forward diffusion process of the DDPM provides necessary information to train a DDPM. It gradually adds noise to existing images $\mathbf{x}_0 \sim p(x)$ using the Ornstein–Uhlenbeck diffusion process (OU process) \citep{Uhlenbeck1930OnTT} within a finite time interval $t\in [0,T]$. The OU process is defined by the stochastic differential equation (SDE):
\begin{equation} \label{1.5 OU process Noise}
d \mathbf{x}_t = - \frac{1}{2} \mathbf{x}_t dt + d\mathbf{W}_t,
\end{equation}
in which $t$ is the forward time of the diffusion process, $\mathbf{x}_t$ is the noise contaminated image at time $t$, and $\mathbf{W}_t$ is a standard Brownian motion. The standard Brownian motion formally satisfies $d \mathbf{W}_t = \sqrt{d t} \boldsymbol{\epsilon}$ with $\boldsymbol{\epsilon}$ be a standard Gaussian noise.
In practice, the OU process is numerically discretized into the variance-preserving (VP) form \citep{Song2020ScoreBasedGM}:
\begin{equation} \label{1.5 forward diffusion process sampling discrete, alt}
   \mathbf{x}_{i} =  \sqrt{1-\beta_{i-1}} \mathbf{x}_{i-1}  + \sqrt{\beta_{i-1}}\boldsymbol{\epsilon}_{i-1},
\end{equation}
where $i = 1,\cdots,n$ is the number of the time step, $\beta_i$ is the step size of each time step, $\mathbf{x}_i$ is  image at $i$th time step with time $t_i = \sum_{j=0}^{i-1} \beta_j$, $\boldsymbol{\epsilon}_{i}$ is standard Gaussian random variable. The time step size usually takes the form $\beta_i = \frac{i(b_2 - b_1)}{n-1} + b_1$ where  $b_1 = 10^{-4}$ and $b_2 = 0.02$. Note that our interpretation of $\beta$ differs from that in \cite{Song2020ScoreBasedGM}, treating $\beta$ as a varying time-step size to solve the autonomous SDE \eqref{1.5 OU process Noise} instead of a time-dependent SDE. Our interpretation holds as long as every $\beta_i^2$ is negligible and greatly simplifies future analysis. The discretized OU process \eqref{1.5 forward diffusion process sampling discrete, alt} adds a small amount of Gaussian noise to the image at each time step $i$, gradually contaminating the image until $\mathbf{x}_n \sim \mathcal{N}(\mathbf{0},I)$.

Training a DDPM aims to recover the original image $x_0$ from one of its contaminated versions $x_i$. In this case \eqref{1.5 forward diffusion process sampling discrete, alt} could be rewritten into the form
\begin{equation} \label{1.5 forward diffusion process sampling discrete, alt ss}
   \mathbf{x}_{i} 
=\sqrt{\bar{\alpha}_i}\mathbf{x}_0 + \sqrt{1-\bar{\alpha}_i}\bar{\boldsymbol{\epsilon}}_i; \quad 1\le i \le n,
\end{equation}
where $\bar{\alpha}_i = \prod_{j=0}^{i-1} (1-\beta_j)$ is the weight of contamination and $\bar{\boldsymbol{\epsilon}}_i$ is a standard Gaussian random noise to be removed. This equation tells us that the distribution of $\mathbf{x}_i$ given $\mathbf{x}_0$ is a Gaussian distribution
\begin{equation} \label{condition distribution App1}
p(\mathbf{x}_i | \mathbf{x}_0) = \mathcal{N}(\mathbf{x}|\sqrt{\bar{\alpha}_i}\mathbf{x}_0, (1-\bar{\alpha}_i) I),
\end{equation}
and the noise $\bar{\boldsymbol{\epsilon}}_i$ is related to the its score function
\begin{equation} \label{score and eps}
\bar{\boldsymbol{\epsilon}}_i = -\sqrt{1-\bar{\alpha}_i}  \mathbf{s}(\mathbf{x}_i | \mathbf{x}_0, t_i),
\end{equation}
where $\mathbf{s}(\mathbf{x}_i | \mathbf{x}_0, t_i)=\nabla_{\mathbf{x}_i} \log p(\mathbf{x}_i | \mathbf{x}_0)$ is the score of the density $p(\mathbf{x}_i | \mathbf{x}_0)$ at $\mathbf{x}_i$. 

DDPM aims to removes the noise $\bar{\boldsymbol{\epsilon}}_i$ from $\mathbf{x}_i$ by training a denoising neural network $\boldsymbol{\epsilon}_\theta( \mathbf{x}, t_i  )$ to predict and remove the noise $\bar{\boldsymbol{\epsilon}}_i $. This means that DDPM minimizes the denoising objective \citep{Ho2020DenoisingDP}:
\begin{equation} \label{1.6 DDPM objective}
\begin{split}
     L_{denoise}(\boldsymbol{\epsilon}_\theta) &=\frac{1}{n}\sum_{i=1}^n \mathbf{E}_{\mathbf{x}_0 \sim p(\mathbf{x})}  \mathbf{E}_{\bar{\boldsymbol{\epsilon}}_i\sim \mathcal{N}(\mathbf{0},I)}\| \bar{\boldsymbol{\epsilon}}_i  -  \boldsymbol{\epsilon}_\theta( \bold{x}_i, t_i  )\|_2^2.
\end{split}
\end{equation}
This is equivalent to, with the help of \eqref{score and eps} and tricks in \citep{Vincent2011ACB}, a denoising score matching objective  
\begin{equation}
\label{1.6 DDPM objective s matching}
\begin{split}
     L_{denoise}(\boldsymbol{\epsilon}_\theta) &=\frac{1}{n}\sum_{i=1}^{n}    (1-\bar{\alpha}_i)\mathbf{E}_{\mathbf{x_i}\sim p(\mathbf{x}_i)} \|  \mathbf{s}(\mathbf{x}_i, t_i)  +  \frac{\boldsymbol{\epsilon}_\theta( \bold{x}_i, t_i  )}{\sqrt{1-\bar{\alpha}_i}}\|_2^2.
\end{split}
\end{equation}
where $ \mathbf{s}(\mathbf{x}, t_i) =\nabla_{\mathbf{x}_i} \log p(\mathbf{x}_i)|_{\mathbf{x}_i = \mathbf{x}}$ is the score function of the density $p(\mathbf{x}_i)$. This objectives says that the denoising neural network $\boldsymbol{\epsilon}_\theta( \mathbf{x}, t_i  )$ is trained to approximate a scaled score function $\boldsymbol{\epsilon}( \mathbf{x}, t_i  )$ \citep{Yang2022DiffusionMA}
\begin{equation} \label{1.6 DDPM result score eps}
\boldsymbol{\epsilon}( \mathbf{x}, t_i  ) = \argmin_{\boldsymbol{\epsilon}_\theta} L(\boldsymbol{\epsilon}_\theta) =  -\sqrt{1-\bar{\alpha}_i}\mathbf{s}(\mathbf{x}, t_i).
\end{equation}
Another useful property we shall exploit later is that for \textbf{infinitesimal} time steps, the contamination weight $\bar{\alpha}_i$ is the exponential of the diffusion time $t_i$
\begin{equation} \label{alpha bar limit}
  \lim_{\max_j \beta_j \xrightarrow[]{}0} \bar{\alpha}_i  \xrightarrow[]{} e^{-t_i}.
\end{equation}
In this case, the discretized OU process \eqref{1.5 forward diffusion process sampling discrete, alt} is equivalent to the OU process \eqref{1.5 OU process Noise}, hence the scaled score function $\boldsymbol{\epsilon}( \mathbf{x}, t_i  )$ is
\begin{equation} \label{1.6 DDPM result score eps, infinitesimal}
\boldsymbol{\epsilon}( \mathbf{x}, t  ) = -\sqrt{1-e^{-t}}\mathbf{s}(\mathbf{x}, t),
\end{equation}
where $\mathbf{s}(\mathbf{x}, t)$ is a solution of the score Fokker-Planck equation \eqref{1.7 OU Fokker Plank of score} of the OU process.

The backward diffusion process is used to sample from the DDPM by removing the noise of an image step by step. It is the time reversed version of the OU process, starting at $x_{0'} \sim \mathcal{N}(\mathbf{x}|\mathbf{0}, I)$, using the reverse of the OU process \citep{Anderson1982ReversetimeDE}:
\begin{equation} \label{1.5 reverse diffusion process}
d\mathbf{x}_{t'} = \left( \frac{1}{2} \mathbf{x}_{t'}+ \mathbf{s}(\mathbf{x}, T-t') \right) dt' + d\mathbf{W}_{t'},
\end{equation} in which $t' \in [0,T]$ is the backward time, $\mathbf{s}(\mathbf{x}, T-t') = \nabla_{\mathbf{x}_{T-t'}} \log p(\mathbf{x}_{T-t'})|_{\mathbf{x}}$ is the score function of the density of $\mathbf{x}_{t=T-t'}$ in the forward process. In practice, the backward diffusion process is discretized into 
\begin{equation} \label{1.5 backward diffusion process sampling discrete, alt}
   \mathbf{x}_{i'+1} = \frac{\mathbf{x}_{i'} + \mathbf{s}(\mathbf{x}_{i'}, T-t'_{i'}) \beta_{n-i'}}{\sqrt{1-\beta_{n-i'}} }  + \sqrt{\beta_{n-i'}}\boldsymbol{\epsilon}_{i'},
\end{equation}
where $i' = 0, \cdots, n$ is the number of the backward time step, $\mathbf{x}_{i'}$ is  image at $i'$th backward time step with time $t_{i'}' = \sum_{j=0}^{i'-1} \beta_{n-1-j} = T - t_{n-i'}$, and  $\mathbf{s}(\mathbf{x}_{i'}, T-t'_{i'})$ is estimated by \eqref{1.6 DDPM result score eps}. This discretization is consistent with \eqref{1.5 reverse diffusion process} as long as $\beta_i^2$ are negligible.

The forward and backward process forms a dual pair when total diffusion time $t_n$ is large enough and the time step sizes are small enough, at which $p(\mathbf{x}_n) = \mathcal{N}(\mathbf{0},I)$. In this case the density of $\mathbf{x}_{t}$ in the forward process and the density of $\mathbf{x}_{t'}$ in the backward process satisfying the relation 
\begin{equation}
    p(\mathbf{x}_{t'})|_{t'=T-t}= p(\mathbf{x}_{t}).
 \end{equation}
This relation tells us that the discrete backward diffusion process generate image samples approximately from the image distribution
\begin{equation}
    p(\mathbf{x}_{n'})\approx p(\mathbf{x}_{0}) =p(\mathbf{x}),
 \end{equation}
despite the discretization error and estimation error of the score function $\mathbf{s}(\mathbf{x}, t')$. An accurate estimation of the score function is one of the key to sample high quality images.

\section{Conditional DDPM and Guidance}\label{App2}

Conditional DDPMs, which generate images based on a given condition $\mathbf{c}$, model the conditional image distribution $p(\mathbf{x} | \mathbf{c})$. One can introduce the dependency on conditions by extending the denoising neural network to include the condition $\mathbf{c}$, represented as $\boldsymbol{\epsilon}(\mathbf{x}|\mathbf{c}, t_i)$. 

Guidance is a technique for conditional image generation that trades off control strength and image diversity. It aims to sample from the distribution \citep{Song2020ScoreBasedGM,Dhariwal2021DiffusionMB}
\begin{equation} \label{classifier guided diffusion p}
    p(\mathbf{x} | \mathbf{c}, \omega) \propto p(\mathbf{c} | \mathbf{x})^{1+\omega}p(\mathbf{x}),
\end{equation}
where $\omega > 0$ is the guidance scale. When $\omega$ is large, guidance control the DDPM to produce samples that have the highest classifier likelihood $p(\mathbf{c} | \mathbf{x})$. 

Another equivalent guidance without the need of the classifier $p(\mathbf{c} | \mathbf{x})$ is the Classifier-free guidance \citep{Ho2022ClassifierFreeDG}:
\begin{equation} \label{guided diffusion p}
    p(\mathbf{x} | \mathbf{c}, \omega) \propto p(\mathbf{x} | \mathbf{c})^{1+\omega}p(\mathbf{x})^{-\omega},
\end{equation}

Sampling from $p(\mathbf{x} | \mathbf{c}, \omega)$ using DDPM requires the corresponding denoising neural network $\boldsymbol{\epsilon}(\mathbf{x}|\mathbf{c}, t_i , \omega)$ that is unknown. Classifier-free guidance provides an approximation of it by linearly combine $\boldsymbol{\epsilon}(\mathbf{x}|\mathbf{c}, t_i )$ and $\boldsymbol{\epsilon}(\mathbf{x}, t_i )$ of conditional and unconditional DDPMs. The classifier-free guidance is inspired by the fact that $\boldsymbol{\epsilon}$ is proportional to the score $\mathbf{s}$ in \eqref{1.6 DDPM result score eps}. At $t = t_0 = 0$, the score of the distribution $p(\mathbf{x} | \mathbf{c}, \omega)$ is the linear combination of scores of $p(\mathbf{x} | \mathbf{c})$ and $p(\mathbf{x})$:
\begin{equation} \label{CFG score decomp}
    \mathbf{s}(\mathbf{x} | \mathbf{c},t_0, \omega ) = (1+\omega) \ \mathbf{s}(\mathbf{x} | \mathbf{c}, t_0)-\omega \ \mathbf{s}(\mathbf{x}, t_0),
\end{equation}
where $ \mathbf{s}(\mathbf{x} | \mathbf{c}, t_0, \omega) = \nabla_{\mathbf{x}}\log  p(\mathbf{x} | \mathbf{c}, \omega) $, $ \mathbf{s}(\mathbf{x} | \mathbf{c}, t_0) = \nabla_{\mathbf{x}}\log  p(\mathbf{x} | \mathbf{c}) $, and $ \mathbf{s}(\mathbf{x},t_0 ) = \nabla_{\mathbf{x}}\log  p(\mathbf{x} ) $. Since scores $\mathbf{s}$ are proportional to de-noising neural netowrks $\boldsymbol{\epsilon}$ according to equation \eqref{1.6 DDPM result score eps}, equation \eqref{CFG score decomp} inspires the classifier free guidance to use the following guided denoising neural network $\boldsymbol{\epsilon}_{CF}(\mathbf{x}|\mathbf{c}, t_i , \omega)$: 
\begin{equation} \label{cfg eps}
    \boldsymbol{\epsilon}_{CF}(\mathbf{x}|\mathbf{c}, t_i , \omega) =(1+\omega) \  \boldsymbol{\epsilon}_\theta(\mathbf{x}|\mathbf{c}, t_i)-\omega \  \boldsymbol{\epsilon}_\theta(\mathbf{x}, t_i),
\end{equation}
where $i$ is the number of time step, $t_i = \sum_{j=0}^{i} \beta_j$, $\mathbf{c}$ is the condition, and $\omega>0$ is the guidance scale. The classifier free guidance exactly computes the de-noising neural network $\boldsymbol{\epsilon}(\mathbf{x}|\mathbf{c}, t_i , \omega)$ of $p(\mathbf{x} | \mathbf{c}, \omega)$ at time $t_0=0$ because of the connection between score function and $\boldsymbol{\epsilon}$ \eqref{1.6 DDPM result score eps}. However, $\boldsymbol{\epsilon}_{CF}$ is not a good approximation of the theoretical $\boldsymbol{\epsilon}(\mathbf{x}|\mathbf{c}, t_i , \omega)$ for most of time steps $0 < i < n$ when the guidance scale \(\omega\) is large.

\section{Fokker-Plank Equation}\label{App3}

The probability density distribution $ p(\mathbf{x}_t)$ of the diffusion process \eqref{1.5 OU process Noise} is a function of $\mathbf{x}$ and $t$, governed by the Fokker-Planck equation of the OU process:
\begin{equation} \label{1.5 OU Fokker Plank}
\frac{\partial p}{\partial t} = \frac{1}{2} \nabla_\mathbf{x} \cdot \left(  \mathbf{x} p \right) + \frac{1}{2} \nabla_\mathbf{x}^2 p.
\end{equation} 
The corresponding score function $\mathbf{s} = \nabla_\mathbf{x} \log p$ is a vector valued function of $\mathbf{x}$ and $t$ governed by the score Fokker-Planck equation:
\begin{equation} \label{1.7 OU Fokker Plank of score App}
\frac{\partial \mathbf{s} }{\partial t} =  \frac{1}{2} \left(    \nabla_\mathbf{x} (\mathbf{s} \cdot \mathbf{x} ) +  \nabla_\mathbf{x}^2  \mathbf{s} + \nabla_\mathbf{x} \|  \mathbf{s} \|^2_2 \right)
\end{equation}
that has been recently studied by \cite{Lai2022FPDiffusionIS} (our score Fokker-Planck equation differs from theirs slightly by noting that $\nabla_\mathbf{x} \nabla_\mathbf{x} \cdot \mathbf{s} = \nabla_\mathbf{x}^2  \mathbf{s}$ where $\mathbf{s}$ is a gradient). This equation holds for both unconditional, conditional, and guided DDPM because they share the same forward diffusion process. Their corresponding initial conditions at time $t=t_0 = 0$  are $\mathbf{s}(\mathbf{x}, t_0)$, $\mathbf{s}(\mathbf{x}|\mathbf{c}, t_0)$ and $\mathbf{s}(\mathbf{x}|\mathbf{c}, t_0, \omega)$. 

However, the score Fokker-Planck equation is a non-linear partial differential equation, which means a linear combination of scores of the unconditional and conditional DDPM is not equivalent to the guided score:
\begin{equation} \label{Fokker planck non-linear}
   \mathbf{s}(\mathbf{x}|\mathbf{c}, t, \omega) \neq (1+\omega) \   \mathbf{s}(\mathbf{x}|\mathbf{c}, t)-\omega \   \mathbf{s}(\mathbf{x}, t); \quad t_0<t<t_n,
\end{equation}
even though their initial conditions at $t=t_0 = 0$ satisfies the linear relation \eqref{CFG score decomp}. Consequently, the classifier-free guidance $\boldsymbol{\epsilon}_{CF}$ is not a good approximation of $\boldsymbol{\epsilon}$ for $p(\mathbf{x} | \mathbf{c}, \omega)$ at $t_0<t<t_n$. Furthermore, $\boldsymbol{\epsilon}_{CF}$ does not correspond to a DDPM of any distribution $p(\mathbf{x})$ when $\omega > 0$, because all DDPMs theoretically have corresponding scores satisfying the score Fokker-Planck \eqref{1.7 OU Fokker Plank of score} while $\boldsymbol{\epsilon}_{CF}$ does not. 

Note that when the diffusion time steps are infinitesimally small, equation \eqref{1.7 OU Fokker Plank of score App} could be rewritten in terms of $\boldsymbol{\epsilon}$ with the help of the relation \eqref{1.6 DDPM result score eps, infinitesimal}.
\begin{equation} \label{1.7 OU Fokker Plank of score App eps}
  \frac{\partial \boldsymbol{\epsilon} }{\partial t}  =  \frac{1}{2} \left(   \nabla_\mathbf{x} (\boldsymbol{\epsilon}\cdot \mathbf{x})  +    \nabla_\mathbf{x}^2 \boldsymbol{\epsilon}  -    \frac{ \nabla_\mathbf{x} \| \boldsymbol{\epsilon}   \|^2_2 }{ \sqrt{1-e^{-t}} }  +  \frac{e^{-t}}{ (1 - e^{-t})} \boldsymbol{\epsilon} \right) 
\end{equation}

Our work aims to address this issue by providing non-linear corrections to the classifier-free guidance, making it approximately satisfy the score Fokker-Planck equation. We propose the Harmonic ansatz which says the Laplacian term $\nabla_\mathbf{x}^2  \mathbf{s}$ in the score Fokker-Plank equation is negligible. It allow us to use the method of characteristics to handle the non-linear term. Noting from \eqref{1.6 DDPM result score eps} that $\boldsymbol{\epsilon} \propto -\mathbf{s}$, the Harmonic ansatz stands as a good approximation as long as $ \| \nabla_{\mathbf{x}}^2  \boldsymbol{\epsilon} \| \ll \|-\nabla_\mathbf{x} ( \boldsymbol{\epsilon} \cdot \mathbf{x} ) + \nabla_\mathbf{x} \|   \boldsymbol{\epsilon} \|^2_2 \|  $ along possible diffusion trajectories of DDPM.

\section{Deriving the Characteristic Guidance Using the Method of Characteristics} \label{App5}

The exactness of the classifier-free guidance revealed by \eqref{CFG score decomp} as initial condition at $t_0=0$ and its failure at $t_i > t_0$ due to non-linearity \eqref{Fokker planck non-linear} are key observations that inspires the characteristic guidance. Characteristic guidance answers the following problem: Given two known solutions $\mathbf{s}_1(\mathbf{x}, t)$ and $\mathbf{s}_2(\mathbf{x}, t)$ of the score Fokker-Planck equation \eqref{1.7 OU Fokker Plank of score}, compute another solution $\mathbf{s}(\mathbf{x}, t)$ of the score Fokker-Planck equation with the following initial condition: 
\begin{equation} \label{APP5 initial condition}
\mathbf{s}(\mathbf{x}, 0) = (1+\omega)\mathbf{s}_1(\mathbf{x}, 0) -\omega \mathbf{s}_2(\mathbf{x}, 0),
\end{equation} 
Note that this initial condition is the same as the condition $
\boldsymbol{\epsilon}(\mathbf{x}, 0) = (1+\omega)\boldsymbol{\epsilon}_1(\mathbf{x}, 0) -\omega \boldsymbol{\epsilon}_2(\mathbf{x}, 0)
$ used in lemma \ref{lemma chara} because score $\mathbf{s}$ is proportional to $\boldsymbol{\epsilon}$ \eqref{1.6 DDPM result score eps}. Similarly, the problem is equivalent to express $\boldsymbol{\epsilon}(\mathbf{x}, t)$ in terms of $\boldsymbol{\epsilon}_1(\mathbf{x}, t)$ and $\boldsymbol{\epsilon}_2(\mathbf{x}, t)$ as stated in lemma \ref{lemma chara}.

It is not easy to express $\mathbf{s}(\mathbf{x}, t)$ as a function of $\mathbf{s}_1(\mathbf{x}, t)$ and $\mathbf{s}_2(\mathbf{x}, t)$ without training or taking complex derivatives in Eq. \eqref{1.7 OU Fokker Plank of score}. Fortunately, the harmonic ansatz \ref{ansatz H} leads us to a viable solution. \textbf{The harmonic ansatz} reduces the score Fokker-Planck equation \eqref{1.7 OU Fokker Plank of score} to the following non-linear first-order partial differential equation (PDE) 
\begin{equation} \label{1.7 OU Fokker Plank of score harmonic}
\frac{\partial \mathbf{s} }{\partial t} = \frac{1}{2} \left( \nabla_\mathbf{x} (\mathbf{s} \cdot \mathbf{x} )+ \nabla_\mathbf{x} \| \mathbf{s} \|^2_2 \right),
\end{equation} 
where the Laplacian term is omitted but the non-linear term remains, capturing the non-linear effect in DDPMs. We treat $\mathbf{s}(\mathbf{x}, t)$, $\mathbf{s}_1(\mathbf{x}, t)$, and $\mathbf{s}_2(\mathbf{x}, t)$ as solutions to this PDE and deduce $\mathbf{s}(\mathbf{x}, t)$ using the method of characteristics.

\paragraph{Characteristic Lines}The method of characteristics reduce a partial differential equation to a family of ordinary differential equations (ODE) by considering solutions along characteristic lines. In our case, we consider the characteristic lines $\mathbf{x}(t)$ satisfying the ODE:
\begin{equation} \label{1.7 OU Fokker Plank of score harmonic charaline}
\frac{d \mathbf{x}(t)}{d t} = -\left( \frac{1}{2} \mathbf{x} + \mathbf{s}(\mathbf{x}(t),t)\right),
\end{equation} 
substitute such characteristic lines into the PDE \eqref{1.7 OU Fokker Plank of score harmonic} yields the dynamics of the score functions along the lines
\begin{equation} \label{1.7 OU Fokker Plank of score harmonic charaline2}
\frac{d \mathbf{s}(\mathbf{x}(t),t)}{d t} = \frac{1}{2} \mathbf{s}(\mathbf{x}(t),t),
\end{equation} 
The characteristic lines of the simplified score Fokker-Planck equation \eqref{1.7 OU Fokker Plank of score harmonic}, which are solutions of the ODEs \eqref{1.7 OU Fokker Plank of score harmonic charaline} and \eqref{1.7 OU Fokker Plank of score harmonic charaline2}, has the form
\begin{equation} \label{App5 charaline}
\begin{split}
       \mathbf{x}(t) &= e^{-\frac{ t}{2} }   \mathbf{x}_0 -    (1- e^{-t }) \mathbf{s}(\mathbf{x}(t),t) \\
       \mathbf{s}(\mathbf{x}(t),t) &= \mathbf{s}(\mathbf{x}_0, 0)e^{\frac{ t}{2} }, \\
\end{split}
\end{equation}
where $\mathbf{x}_0 = \mathbf{x}(0)$ is the starting point of the line and serves as a constant parameter, $\mathbf{s}(\mathbf{x}_0, 0)$ depends on the initial condition of $\mathbf{s}$ at $t=0$. Besides, such characteristic lines of $\mathbf{s}$ also induces characteristic lines of $\boldsymbol{\epsilon}$ as they are connected by \eqref{1.6 DDPM result score eps, infinitesimal}.

An important remark on the characteristic lines is that \eqref{App5 charaline} is valid for $\mathbf{s}_{1}$ and $\mathbf{s}_{2}$ as well, if we replace every $\mathbf{s}$ with $\mathbf{s}_{1}$ or $\mathbf{s}_{2}$. This is because $\mathbf{s}$, $\mathbf{s}_{1}$, and $\mathbf{s}_{2}$ are solutions to the same PDE \eqref{1.7 OU Fokker Plank of score harmonic}. The merit of these characteristic lines lies in their ability to determine the value of $ \mathbf{s}(\mathbf{x}(t),t)$ at $t > 0$ using information of $\mathbf{x}_0$ and $\mathbf{s}(\mathbf{x}_0,0)$ at $t=0$, thereby allowing us to utilize the initial condition \eqref{APP5 initial condition}.

\paragraph{Deducing $\mathbf{s}(\mathbf{x},t)$}Our next move is deduce the value $\mathbf{s}(\mathbf{x},t)$ from $\mathbf{s}_1(\mathbf{x},t)$ and $\mathbf{s}_2(\mathbf{x},t)$ with the help of characteristic lines. Denote the characteristic lines of the three score functions $\mathbf{s}$, $\mathbf{s}_1$, and $\mathbf{s}_2$ to be $\mathbf{x}(t)$, $\mathbf{x}_1(t)$, $\mathbf{x}_2(t)$. If the three characteristic lines meets at $\mathbf{x}_0$ at $t=0$
\begin{equation}
    \mathbf{x}(0) =  \mathbf{x}_1(0) =  \mathbf{x}_2(0) = \mathbf{x}_0,
\end{equation}
then the initial condition \eqref{APP5 initial condition} and the equation \eqref{App5 charaline} tells that
\begin{equation} \label{APP5 initial condition match}
\mathbf{s}(\mathbf{x}(t), t) = (1+\omega)\mathbf{s}_1(\mathbf{x}_1(t), t) -\omega \mathbf{s}_2(\mathbf{x}_2(t), t),
\end{equation} 
which achieves our goal in expressing the value $\mathbf{s}$ in terms of $\mathbf{s}_1$ and $\mathbf{s}_2$. In summary, the following system of equations determines the value of $\mathbf{s}$ in terms of $\mathbf{s}_1$ and $\mathbf{s}_2$
\begin{equation} \label{App5 sys4}
\left\{
\begin{array}{rcl}
 \mathbf{x}(t) e^{\frac{ t}{2} } + (e^{\frac{ t}{2} }- e^{-\frac{ t}{2} }) \mathbf{s}(\mathbf{x}(t),t) &=&   \mathbf{x}_0   \\
 \mathbf{x}_1(t) e^{\frac{ t}{2} } + (e^{\frac{ t}{2} }- e^{-\frac{ t}{2} }) \mathbf{s}_1(\mathbf{x}_1(t),t) &=&   \mathbf{x}_0   \\
 \mathbf{x}_2(t) e^{\frac{ t}{2} } + (e^{\frac{ t}{2} }- e^{-\frac{ t}{2} }) \mathbf{s}_2(\mathbf{x}_2(t),t) &=&   \mathbf{x}_0   \\
(1+\omega)\mathbf{s}_1(\mathbf{x}_1(t), t) -\omega \mathbf{s}_2(\mathbf{x}_2(t), t) &=&  \mathbf{s}(\mathbf{x}(t), t),
\end{array}
\right.
\end{equation}
in which the first three equations are reformulations of \eqref{App5 charaline} and say that the three characteristic lines meets at the same $\mathbf{x}_0$ at $t=0$. In practice, we wish to compute the value of $\mathbf{s}$ for a given $\mathbf{x}(t)$ at time $t$, with the functions $\mathbf{s}_1$ and $\mathbf{s}_2$ known in advance. This completes the system with four equations and four unknowns: $\mathbf{x}_0$, $\mathbf{x}_1(t)$, $\mathbf{x}_2(t)$, and $\mathbf{s}(\mathbf{x}(t),t)$.

Since the dependency of $\mathbf{x}$ on $t$ is already characterized by the system of equation, we can simplify the notation by omitting it in the following discussion. Hence, we write $\mathbf{x}$,  $\mathbf{x}_1$, $\mathbf{x}_2$ instead of $\mathbf{x}(t)$,  $\mathbf{x}_1(t)$, $\mathbf{x}_2(t)$ in the following discussion.
\paragraph{Simplify the System} It is possible to simplify the system of four equations into one. We start from eliminating the unknowns $\mathbf{s}(\mathbf{x},t)$ and $\mathbf{x}_0$ with some algebra, reducing the system into two equations
\begin{equation} \label{App5 sys2}
\left\{
\begin{array}{rcl}
 \mathbf{x}_1  &=&  \mathbf{x} + \omega \left( \mathbf{s}_1(\mathbf{x}_1,t) - \mathbf{s}_2(\mathbf{x}_2,t)  \right) (1-e^{-t}) \\
\mathbf{x}_2  &=&  \mathbf{x} + (1+\omega) \left( \mathbf{s}_1(\mathbf{x}_1,t) - \mathbf{s}_2(\mathbf{x}_2,t)  \right) (1-e^{-t}).
\end{array}
\right.
\end{equation}
The above equations also indicate that $\mathbf{x}_1$ and  $\mathbf{x}_2$ are not independent but linearly correlated as
\begin{equation} \label{App5 x12}
     (1+\omega)\mathbf{x}_1- \omega \mathbf{x}_2 = \mathbf{x}.
\end{equation}
This inspires us to define the correction term $\Delta \mathbf{x}$ as
\begin{equation} \label{App5 x12 def}
\begin{split}
    \mathbf{x}_1 &= \mathbf{x} + \omega \Delta \mathbf{x} \\
    \mathbf{x}_2 &= \mathbf{x} + (1+\omega) \Delta \mathbf{x},
\end{split}
\end{equation}
which is exactly the same notation used in the characteristic guidance \eqref{Method characteristic guidance eps}. 

Finally, by combining \eqref{App5 sys2} and \eqref{App5 x12}, we obtain a single equation for $\Delta \mathbf{x}$ as follows:
\begin{equation}\label{App5 delta x in score}
\Delta \mathbf{x} = \left( \mathbf{s}_1(\mathbf{x} + \omega \Delta \mathbf{x},t) - \mathbf{s}_2(\mathbf{x} + (1+\omega) \Delta \mathbf{x},t)  \right) (1-e^{-t}).
\end{equation}
Once we solve for $\Delta \mathbf{x} $ from the above equation, we can easily compute the desired value of the score function $\mathbf{s}(\mathbf{x}, t)$ by using \eqref{App5 x12 def} and \eqref{App5 sys4}:
\begin{equation} \label{score addition App5}
\mathbf{s}(\mathbf{x}, t) = (1+\omega)\mathbf{s}_1(\mathbf{x}_1, t) -\omega \mathbf{s}_2(\mathbf{x}_2, t).
\end{equation}
\paragraph{Characteristic Guidance, Lemma \ref{lemma chara}, and Theorem \ref{th chara}} Now we have nearly completed the derivation of the characteristic guidance. It remains to replace the score functions $\mathbf{s}$ with the de-noising neural networks $\boldsymbol{\epsilon}$ used in DDPM. They are related by \eqref{1.6 DDPM result score eps, infinitesimal} if the diffusion time steps in DDPM are small enough. Consequently we rewrite \eqref{score addition App5} and \eqref{App5 delta x in score} into
\begin{equation} \label{App5 eps ++}
\boldsymbol{\epsilon}(\mathbf{x}, t) = (1+\omega)\boldsymbol{\epsilon}_1(\mathbf{x}_1, t) -\omega \boldsymbol{\epsilon}_2(\mathbf{x}_2, t),
\end{equation}
and
\begin{equation}
\Delta \mathbf{x} = \left(  \boldsymbol{\epsilon}_2(\mathbf{x}_2,t) - \boldsymbol{\epsilon}_1(\mathbf{x}_1,t) \right)\sigma(t),
\end{equation}
where $\sigma(t) = \sqrt{1-e^{-t}}$. These equations are exactly the content of the Lemma \ref{lemma chara}. 

In practice, DDPMs do not use infinitesimal time steps. Therefore we replace $e^{-t_i}$ with $\bar{\alpha}_i$ everywhere since they are related through \eqref{alpha bar limit}. This leads to the characteristic guidance \eqref{Method characteristic guidance eps} and \eqref{non-linear delta relation}.

As for Theorem \ref{th chara}, it is a direct consequence of Lemma \ref{lemma chara}. Lemma \ref{lemma chara} says that $\mathbf{s}(\mathbf{x}, t)$ is a solution of the equation \eqref{1.7 OU Fokker Plank of score harmonic}, then its corresponding $\boldsymbol{\epsilon}(\mathbf{x}, t) $, automatically satisfies the FP equation \eqref{1.7 OU Fokker Plank of score} except the Laplacian term, which can be further eliminated by the Harmonic ansatz. Therefore $\boldsymbol{\epsilon}(\mathbf{x}, t) $ in \eqref{App5 eps ++} has no mixing error \eqref{Background mixing error} under the Harmonic ansatz.

\section{Fixed Point Iteration Methods for Solving the Correction Term} \label{App4}

\begin{algorithm}
\caption{Successive over-relaxation iteration for $\Delta \mathbf{x}$}\label{SOR ite}
\begin{algorithmic}[1]
\REQUIRE $\mathbf{x}, \mathbf{c}, t_i, \omega, \boldsymbol{\epsilon}_\theta, \sigma_i$, $\eta$(tolerance)
\REQUIRE $\mathbf{x}, \mathbf{c}, t_i, \omega, \sigma_i $ \COMMENT{Inputs for the DDPM model}
\REQUIRE $\gamma$ (lr) \COMMENT{Standard parameters for gradient descent}
\REQUIRE $\eta$ \COMMENT{The tolerance as stopping criteria}
\ENSURE $\Delta \mathbf{x}$
\STATE Initialize $\Delta \mathbf{x}^{(0)}$ as a zero vector
\STATE Set $k=0$
\REPEAT
\STATE $k \gets k+1$
\STATE $\mathbf{x}_1^{(k-1)} = \mathbf{x} + \omega  \Delta \mathbf{x}^{(k-1)}$
\STATE $\mathbf{x}_2^{(k-1)} = \mathbf{x}+ (1+\omega)\Delta \mathbf{x}^{(k-1)}$
\STATE $\mathbf{g}^{(k)} = \Delta \mathbf{x}^{(k-1)} -\mathbf{P} \circ \left( \boldsymbol{\epsilon}_\theta( \mathbf{x}_2^{(k-1)}, t_i  ) -\boldsymbol{\epsilon}_\theta(\mathbf{x}_1^{(k-1)} |\mathbf{c}, t_i  ) \right) \sigma_i, $  
\STATE Update $\Delta \mathbf{x}^{(k)} =  \Delta \mathbf{x}^{(k-1)}  - \gamma \mathbf{g}^{(k)}  $, 
\UNTIL $\| \mathbf{g} \|_2^2 < \eta^2 \mbox{dim}(\mathbf{g})  $
\STATE Return $\Delta \mathbf{x}^{(k)}$ as $\Delta \mathbf{x}$
\end{algorithmic}
\end{algorithm}

\begin{algorithm}
\caption{RMSprop iteration for $\Delta \mathbf{x}$}\label{RMS ite}
\begin{algorithmic}[1]
\REQUIRE $\mathbf{x}, \mathbf{c}, t_i, \omega, \sigma_i $ \COMMENT{Inputs for the DDPM model}
\REQUIRE $\alpha$, $\epsilon$, $\gamma$ (lr) \COMMENT{Standard parameters for the RMSprop algorithm}
\REQUIRE $\eta$ \COMMENT{The tolerance as stopping criteria}
\REQUIRE $D$ \COMMENT{Learning rate schedule}
\ENSURE $\Delta \mathbf{x}$
\STATE Initialize $\Delta \mathbf{x}^{(0)}$ as a zero vector
\STATE Set $k=0$
\REPEAT
\STATE $k \gets k+1$
\STATE $\mathbf{x}_1^{(k-1)} = \mathbf{x} + \omega  \Delta \mathbf{x}^{(k-1)}$
\STATE $\mathbf{x}_2^{(k-1)} = \mathbf{x}+ (1+\omega)\Delta \mathbf{x}^{(k-1)}$
\STATE $\mathbf{g}^{(k)} = \Delta \mathbf{x}^{(k-1)} -\mathbf{P} \circ \left( \boldsymbol{\epsilon}_\theta( \mathbf{x}_2^{(k-1)}, t_i  ) -\boldsymbol{\epsilon}_\theta(\mathbf{x}_1^{(k-1)} |\mathbf{c}, t_i  ) \right) \sigma_i, $  
\STATE $\mathbf{v}^{(k)} = \alpha \mathbf{v}^{(k-1)} + (1-\alpha )( \mathbf{g}^{(k)})^2$ 
\STATE $\gamma^{(k)}$ = $\gamma/(1 + D k)$  
\STATE Update $\Delta \mathbf{x}^{(k)} =  \Delta \mathbf{x}^{(k-1)}  - \gamma^{(k)} \mathbf{g}^{(k)} / ( \sqrt{ \mathbf{v}^{(k)}}+\epsilon) $, 
\UNTIL $\| \mathbf{g} \|_2^2 < \eta^2 \mbox{dim}(\mathbf{g})  $
\STATE Return $\Delta \mathbf{x}^{(k)}$ as $\Delta \mathbf{x}$
\end{algorithmic}
\end{algorithm}
We solve $\Delta \mathbf{x}$ from the equation \eqref{non-linear delta relation} by fixed point iterative methods. The key idea is treating the residual vector $\mathbf{g}$ of the equation \eqref{non-linear delta relation}
$$\mathbf{g} = \Delta \mathbf{x} -\mathbf{P} \circ \left( \boldsymbol{\epsilon}_\theta( \mathbf{x}_2, t_i  ) -\boldsymbol{\epsilon}_\theta(\mathbf{x}_1 |\mathbf{c}, t_i  ) \right) \sigma_i, $$  
as the gradient in an optimization problem. Consequently we could apply any off-the-shelf optimization algorithm to solve the fixed point iteration problem. 

For example, adopting the gradient descent methods in  leads us to the well-known successive over-relaxation iteration method for fixed point problem (Alg.\ref{SOR ite}). Similarly, adopting the RMSprop method leads us to Alg.\ref{RMS ite} with faster convergence.

Another kind of effective iteration method to solve $\Delta \mathbf{x}$ is the Anderson acceleration method \citep{Walker2011AndersonAF} Alg.\ref{AA ite}, which turns out to be faster than RMSprop when applied to latent diffusion models. 
\begin{algorithm}
\caption{Anderson acceleration (AA) iteration for $\Delta \mathbf{x}$}\label{AA ite}
\begin{algorithmic}[1]
\REQUIRE $\mathbf{x}, \mathbf{c}, t_i, \omega, \sigma_i $ \COMMENT{Inputs for the DDPM model}
\REQUIRE $m \ge 2$, $\gamma$ (lr), $\Delta \mathbf{x}_B = [\ ]$, $\mathbf{g}_B = [\ ]$ \COMMENT{Parameters and buffers for AA algorithm}
\REQUIRE $\eta$ \COMMENT{The tolerance as stopping criteria}
\ENSURE $\Delta \mathbf{x}$
\STATE Initialize $\Delta \mathbf{x}^{(0)}$ as a zero vector
\STATE Set $k=0$
\REPEAT
\STATE $k \gets k+1$
\STATE $\mathbf{x}_1^{(k-1)} = \mathbf{x} + \omega  \Delta \mathbf{x}^{(k-1)}$
\STATE $\mathbf{x}_2^{(k-1)} = \mathbf{x}+ (1+\omega)\Delta \mathbf{x}^{(k-1)}$
\STATE $\mathbf{g}^{(k)} = \Delta \mathbf{x}^{(k-1)} -\mathbf{P} \circ \left( \boldsymbol{\epsilon}_\theta( \mathbf{x}_2^{(k-1)}, t_i  ) -\boldsymbol{\epsilon}_\theta(\mathbf{x}_1^{(k-1)} |\mathbf{c}, t_i  ) \right) \sigma_i, $  
\STATE $\Delta \mathbf{x}_B$.append($\Delta \mathbf{x}^{(k-1)}$), \ $\mathbf{g}_B$.append($\mathbf{g}^{(k)}$)
\IF{$\mbox{len}(\Delta \mathbf{x}_B) \ge 2$}
    \STATE $\mathbf{g}_B[-2] = \mathbf{g}_B[-1] - \mathbf{g}_B[-2]$
    \STATE $\Delta \mathbf{x}_B[-2] = \Delta \mathbf{x}_B[-1] - \Delta \mathbf{x}_B[-2]$
    \IF{$\mbox{len}(\Delta \mathbf{x}_B) > m$}
        \STATE $\mbox{delete}\ \mathbf{g}_B[0]$; \ $\mbox{delete} \ \Delta \mathbf{x}_B[0]$; 
    \ENDIF
    \STATE $A_g = \mathbf{g}_B[:-1]$; $A_x = \Delta \mathbf{x}_B[:-1]$;
    \STATE $\mathbf{b}_g = \mathbf{g}_B[-1]$; $\mathbf{b}_x = \Delta \mathbf{x}_B[-1]$;
    \STATE $\mathbf{w}  =  \argmin_{\mathbf{w}} \|  A_g \mathbf{w} - \mathbf{b}_g \|_2^2$
    \STATE $ \Delta \mathbf{x}_{AA}^{(k-1)} = \mathbf{b}_x - A_x \mathbf{w}$
    \STATE $ \mathbf{g}^{(k)}_{AA} = \mathbf{b}_g - A_g \mathbf{w}$
\ELSE
    \STATE $ \Delta \mathbf{x}_{AA}^{(k-1)} = \Delta \mathbf{x}^{(k-1)}$
    \STATE $ \mathbf{g}^{(k)}_{AA} = \mathbf{g}^{(k)}$ 
\ENDIF
\STATE Update $\Delta \mathbf{x}^{(k)} =  \Delta \mathbf{x}^{(k-1)}_{AA}  - \gamma \mathbf{g}^{(k)}_{AA}$, 
\UNTIL $\| \mathbf{g} \|_2^2 < \eta^2 \mbox{dim}(\mathbf{g})  $
\STATE Return $\Delta \mathbf{x}^{(k)}$ as $\Delta \mathbf{x}$
\end{algorithmic}
\end{algorithm}

In practice, the number of iterations required to solve $\Delta \mathbf{x}$ depends on both the iteration method and the DDPM model. Different iteration methods may have different convergence rates and stability properties. Therefore, we suggest trying different iteration methods to find the most efficient and effective one for a given DDPM model.

\section{Conditional Gaussian} \label{App6}

This experiment compares the classifier free guidance and the characteristic guidance on sampling from a conditional 2D Gaussian distributions, which strictly satisfies the harmonic ansatz. 

The diffusion model models two distribution: the conditional distribution $p(x_1,x_2|c_1,c_2)$ and the unconditional distribution $p(x_1,x_2)$. In our case, both of them are 2D Gaussian distributions
\begin{equation} 
\begin{split}
p(x_1,x_2|c_1,c_2)&=\mathcal{N}\left( x_1,x_2 |(c_1,c_2)^T,  I \right)\\  
p(x_1,x_2)&=\mathcal{N}\left( x_1,x_2 |(0,0)^T, 5 I \right) 
\end{split}
\end{equation}
The guided diffusion model \eqref{guided diffusion p} aims to sample from the distribution 
\begin{equation} \label{Gaussian guided diffusion analytical}
p(x_1,x_2|c_1,c_2;\omega)=\mathcal{N}\left(x_1,x_2 | (\frac{5 c_1 (\omega+1)}{4\omega+5},\frac{5 c_2 (\omega+1) }{4\omega+5})^T, \frac{5}{4\omega+5} I \right) 
\end{equation}
having the properties $p(x_1,x_2|c_1,c_2;0) = p(x_1,x_2|c_1,c_2)$ and $p(x_1,x_2|c_1,c_2;-1) = p(x_1,x_2)$.

The score function of the unconditional, conditional, and guided model along the forward diffusion process  can be solved analytically from the FP equation:
$$\mathbf{s}(x_1,x_2|c_1,c_2; t, \omega)=\left(\frac{5 c_1 (\omega+1) e^{t/2}-(4\omega+5) e^t x_1}{(4\omega+5) e^t-4 \omega},\frac{5 c_2 (\omega+1) e^{t/2}-(4\omega+5) e^t x_2}{(4\omega+5) e^t-4 \omega }\right)^T$$
with $\omega = -1$ and $\omega = 0$ corresponds to the scores of unconditional and conditional distributions. Therefore the denoising neural networks $\boldsymbol{\epsilon}_\theta$ of unconditional and conditional DDPMs could be analytically calculated as
\begin{equation} \label{gaussian eps}
\begin{split}
\boldsymbol{\epsilon}_\theta( x_1,x_2 , t  ) &=  -\sqrt{1-e^{-t}}\mathbf{s}(x_1,x_2;t,-1)\\
\boldsymbol{\epsilon}_\theta( x_1,x_2|c_1, c_2; t  ) &=  -\sqrt{1-e^{-t}}\mathbf{s}(x_1,x_2|c_1,c_2; t, 0)\\
\end{split}
\end{equation}
The corresponding classifier free guided $\boldsymbol{\epsilon}_{CF}$ and characteristic guided $\boldsymbol{\epsilon}_{CH}$ are computed according to \eqref{cfg eps} and \eqref{Method characteristic guidance eps}. The characteristic guidance is computed with Alg.\ref{RMS ite} with $\gamma=0.01$, $\eta = 0.01$, and $\alpha = 0.9999$.

\begin{figure}[h]
\begin{center}
\includegraphics[scale=0.45]{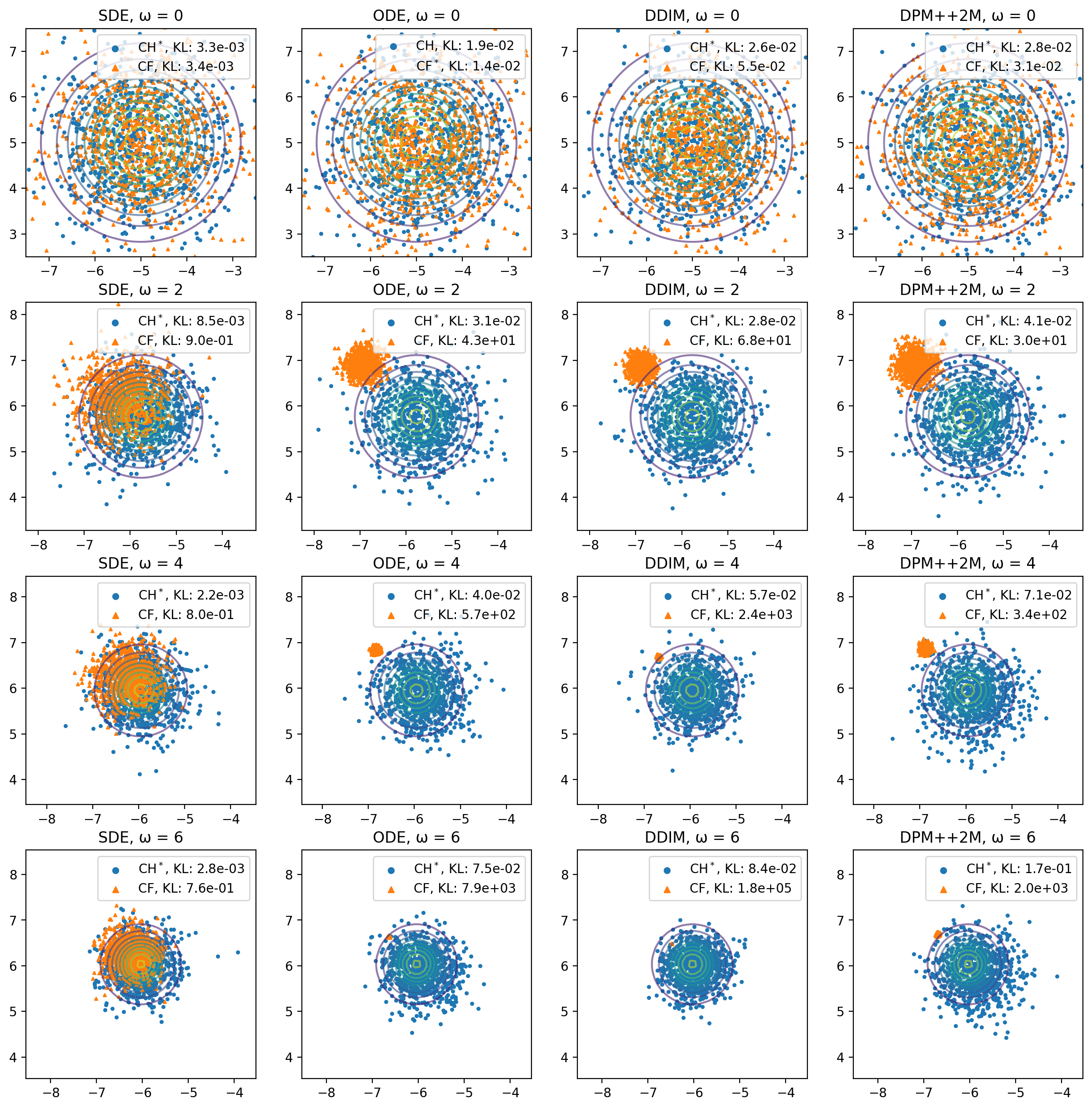}
\end{center}
\caption{Comparison between characteristic guidance (CH) and classifier free guidance (CF) of DDPM on the conditional Gaussian experiment. Samples are drawn with various sample methods: SDE, probabilistic ODE, DDIM and DPM++2M. The contours corresponds to the theoretical reference distribution of guided DDPMs in \eqref{Gaussian guided diffusion analytical}. } \label{The sampling results Gaussian}
\end{figure}

We sample from guided DDPMs using four different ways: the SDE \eqref{1.5 backward diffusion process sampling discrete, alt}, probabilistic ODE \cite{Song2020ScoreBasedGM}, DDIM \cite{Song2020DenoisingDI}, and DPM++2M \cite{Lu2022DPMSolverFS}. We set $c_1 = -5$, $c_2 = 5$,  $\beta_1 = 1e-4$,
$\beta_2 = 0.015$ and the total time step to be $n=1000$ for SDE and ODE, and $n=20$ for DDIM and DPM++2M. The sample results are compared with the theoretical reference distribution \eqref{Gaussian guided diffusion analytical} by assuming the samples are Gaussian distributed then computing the KL divergence with the theoretical reference. The results are plotted in Fig.\ref{The sampling results Gaussian}. Our characteristic guidance outperforms classifier free guidance in all cases with $\omega>0$.

\section{Mixture of Gaussian} \label{App7}

This experiment compares the classifier free guidance and the characteristic guidance when the assumption $\nabla_\mathbf{x} \nabla_\mathbf{x} \cdot \mathbf{s} = 0$ is not satisfied. We train conditional and unconditional DDPM to learn two distributions: the conditional distribution $p(\mathbf{x}|\mathbf{c})$ and the unconditional distribution $p(\mathbf{x}) = p(\mathbf{x})$.
\begin{equation} \label{MGaussian guided diffusion analytical}
\begin{split}
p(\mathbf{x}|\mathbf{c})&=\mathcal{N}\left(\mathbf{x}|(-1,-1/\sqrt{3})^T,  I \right)^{c_0} \mathcal{N}\left(\mathbf{x}|(1,-1/\sqrt{3})^T,  I \right)^{c_1} \mathcal{N}\left(\mathbf{x}|(0,\sqrt{3} -1/\sqrt{3})^T,  I \right)^{c_2} \\  
p(\mathbf{x})&=\left(p(\mathbf{x}|(1,0,0)^T) + p(\mathbf{x}|(0,1,0)^T) + p(\mathbf{x}|(0,0,1)^T) \right)/3
\end{split}
\end{equation}
where $\mathbf{c}$ is a three dimensional one-hot vector.

The guided diffusion model aims to sample from the distribution 
\begin{equation} \label{MGaussian guided diffusion analytical}
p(\mathbf{x}|\mathbf{c};\omega)= \frac{1}{Z(\omega, \mathbf{c})} p(\mathbf{x} | \mathbf{c})^{1+\omega}p(\mathbf{x})^{-\omega}
\end{equation}
where $Z(\omega, \mathbf{c}) = \int p(\mathbf{x} | \mathbf{c})^{1+\omega}p(\mathbf{x})^{-\omega} d \mathbf{x}$ is the partition function that can be computed with the Monte Carlo method numerically. 

We train the denoising neural networks $\boldsymbol{\epsilon}_\theta(\mathbf{x}|\mathbf{c},t)$ and $\boldsymbol{\epsilon}_\theta(\mathbf{x},t)$ for conditional and unconditional neural network on a dataset of 20000 pairs of $(\mathbf{x}, \mathbf{c})$. Particularly, The $\boldsymbol{\epsilon}_\theta(\mathbf{x},t)$ is trained as a special case of conditioned $\boldsymbol{\epsilon}_\theta(\mathbf{x}|\mathbf{c},t)$ with $\mathbf{c} = \mathbf{0}$. The characteristic guidance is computed by Alg.\ref{RMS ite} with $\gamma=0.05$, $\eta = 0.02$, and $\alpha = 0.99$.

\begin{figure}[h]
\begin{center}
\includegraphics[width=0.8\columnwidth]{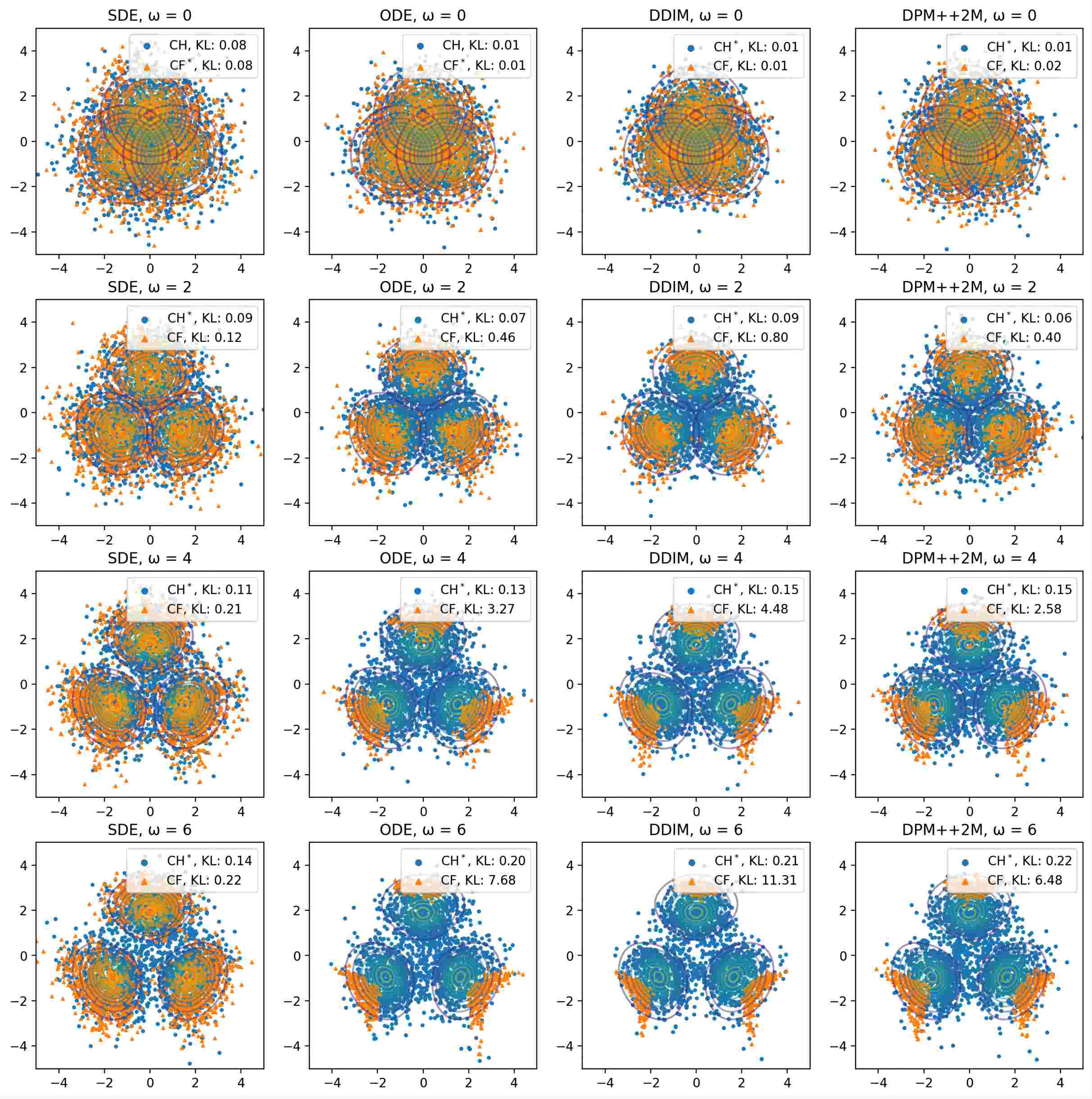}
\end{center}
\caption{Comparison between characteristic guidance (CH) and classifier free guidance (CF) of DDPM on the mixture of Gaussian experiment. Samples are drawn with various sample methods: SDE, probabilistic ODE, DDIM and DPM++2M. The contours corresponds to the theoretical reference distribution of guided DDPMs in \eqref{MGaussian guided diffusion analytical}. } \label{The sampling results M Gaussian}
\end{figure}

We sample from guided DDPMs using four different ways: the SDE \eqref{1.5 backward diffusion process sampling discrete, alt}, probabilistic ODE \cite{Song2020ScoreBasedGM}, DDIM \cite{Song2020DenoisingDI}, and DPM++2M \cite{Lu2022DPMSolverFS}. During sampling, we set $\beta_1 = 1e-4$,
$\beta_2 = 0.02$ and the total time step to be $n=500$ for SDE and ODE, and $n=20$ for DDIM and DPM++2M. The sample results are compared with the theoretical reference distribution \eqref{MGaussian guided diffusion analytical} by assuming the samples from DDPM are Gaussian distributed then computing the KL divergence with the theoretical reference. The results are plotted in Fig.\ref{The sampling results M Gaussian}. Our characteristic guidance outperforms classifier free guidance in all cases with $\omega>0$.

Fig.\ref{The sampling results M Gaussian} also shows a significant loss of diversity for classifier free guided ODE samplers when $\omega>0$. This loss of diversity is corrected by our characteristic guidance, yielding samples that perfectly matches the theoretical reference.

\section{Diffusion Model for Science: Cooling of Magnet} \label{App8}

Guidance, as described in \ref{guided diffusion p}, can be conceptualized as a cooling process with the guidance scale $\omega$ representing the temperature drop. We demonstrate that the guidance can simulate the cooling process of a magnet around the Curie temperature, at which a paramagnetic material gains permanent magnetism through phase transition. We also demonstrate that non-linear correction for classifier-free guidance is important for correctly characterize the megnet's phase transition. 

The phase transition at the Curie temperature can be qualitatively described by a scalar Landau-Ginzburg model. We consider a thin sheet of 2D magnetic material which can be magnetized only in the direction perpendicular to it. The magnetization of this material at each point is described by a scalar field $\phi(\mathbf{x})$ specifying the magnitude and the direction of magnetization. When $\phi$ has a higher absolute value, it means that region is more strongly magnetized. Numerically, we discretize $\phi(\mathbf{x})$ into periodic $8\times8$ grid, with $i$th grid node assigned with a float number $\phi_{i}$. It is equivalent to say that the magnetization field $\phi(\mathbf{x})$ is represented by $8\times8$ picture with $1$ channel. At any instance, there is a field $\phi$ describing the current magnetization of our magnet.

Thermal movements of molecules in our magnet gradually change our magnet's magnetization $\phi$ from one configuration to another. If we keep track of the field $\phi$ every second for a long duration, the probability of observing a particular configuration of $\phi$ is proportional to the Boltzmann distribution
\begin{equation} \label{Boltzmann distribution}
    p(\phi;T) \propto e^{-\beta H(\phi;T)},
\end{equation}
where $\beta H(\phi;T)$ is Hamiltonian at temperature $T$, assigning an energy to each of possible $\phi$. Around the Curie temperature, the Landau-Ginzburg model of our magnet use the following Hamiltonian:
\begin{equation} \label{LG hamilton}
   \beta H(\phi; T) =  K\left( \frac{1}{2} \sum_{\langle i, j \rangle} (\phi_i - \phi_j)^2 + \sum_i \left( \frac{m^2}{2} (T - T_c) \phi_i^2 + \frac{\lambda}{4!} \phi_i^4 \right) \right) 
\end{equation}
where $m^2=0.1$, $\lambda=1.0$, $K=1$ are parameters and $T_c=200$ is the Curie temperature. The first term sums over adjacent points $i$ and $j$ on the grid, representing the interaction between neighbors. The second term describes self-interaction of each grid. The Landau-Ginzburg model allow us to control the temperature $T$ of magnet using guidance similar to \eqref{CFG score decomp}
$$
\boldsymbol{s}(\phi; (1+\omega) T_1 - \omega T_0) =  (1+\omega)  \boldsymbol{s}(\phi;T_1)-\omega \boldsymbol{s}(\phi;T_0)
$$
where $\omega>0$ is the guidance scale, $T_0 > T_1$ are two distinct temperature, and $\boldsymbol{s}(\phi;T) = \nabla \log P(\phi;T)$ is the score of the Boltzmann distribution in \eqref{Boltzmann distribution}. This means training DDPMs of the magnetization field $\phi$ at two distinct temperature $T_0 > T_1$ theoretically allow us to sample $\phi$ at temperatures below $T_1$ using guidance, corresponding to the cooling of the magnet.

To simulate the cooling of the magnet, we train a conditional DDPM of $\phi$ for two distinct temperatures $p(\phi|T_0=201)$ and $p(\phi|T_1=200)$. The dataset consists of 60000 samples at $T_0=201$ and 60000 samples at $T_1=200$ generated by the Metropolis-Hastings algorithm. Then we sample from the trained DDPM using classifier-free guidance in \eqref{cfg eps} and characteristic guidance in \eqref{Method characteristic guidance eps}
\begin{equation} \label{cfg eps phi4}
\begin{split}
    \boldsymbol{\epsilon}_{CF}(\mathbf{x}|(1+\omega) T_1 - \omega T_0, t_i) &=(1+\omega) \  \boldsymbol{\epsilon}_\theta(\mathbf{x}|T_1, t_i)-\omega \  \boldsymbol{\epsilon}_\theta(\mathbf{x}|T_0, t_i)\\
    \boldsymbol{\epsilon}_{CH}(\mathbf{x}|(1+\omega) T_1 - \omega T_0, t_i) &= (1+\omega) \  \boldsymbol{\epsilon}_\theta( \mathbf{x} + \omega  \Delta \mathbf{x}| T_1, t_i )-\omega \  \boldsymbol{\epsilon}_\theta(\mathbf{x}+ (1+\omega) \Delta \mathbf{x}|  T_0, t_i).
\end{split}
\end{equation}
The characteristic guidance is computed with Alg.\ref{RMS ite} with $\gamma=0.01$, $\eta = 0.1$, and $\alpha = 0.999$. Samples from the guided DDPM is treated approximately as samples at temperature $T=(1+\omega) T_1 - \omega T_0$.

\begin{figure}[h]
\begin{center}
\includegraphics[scale=0.35]{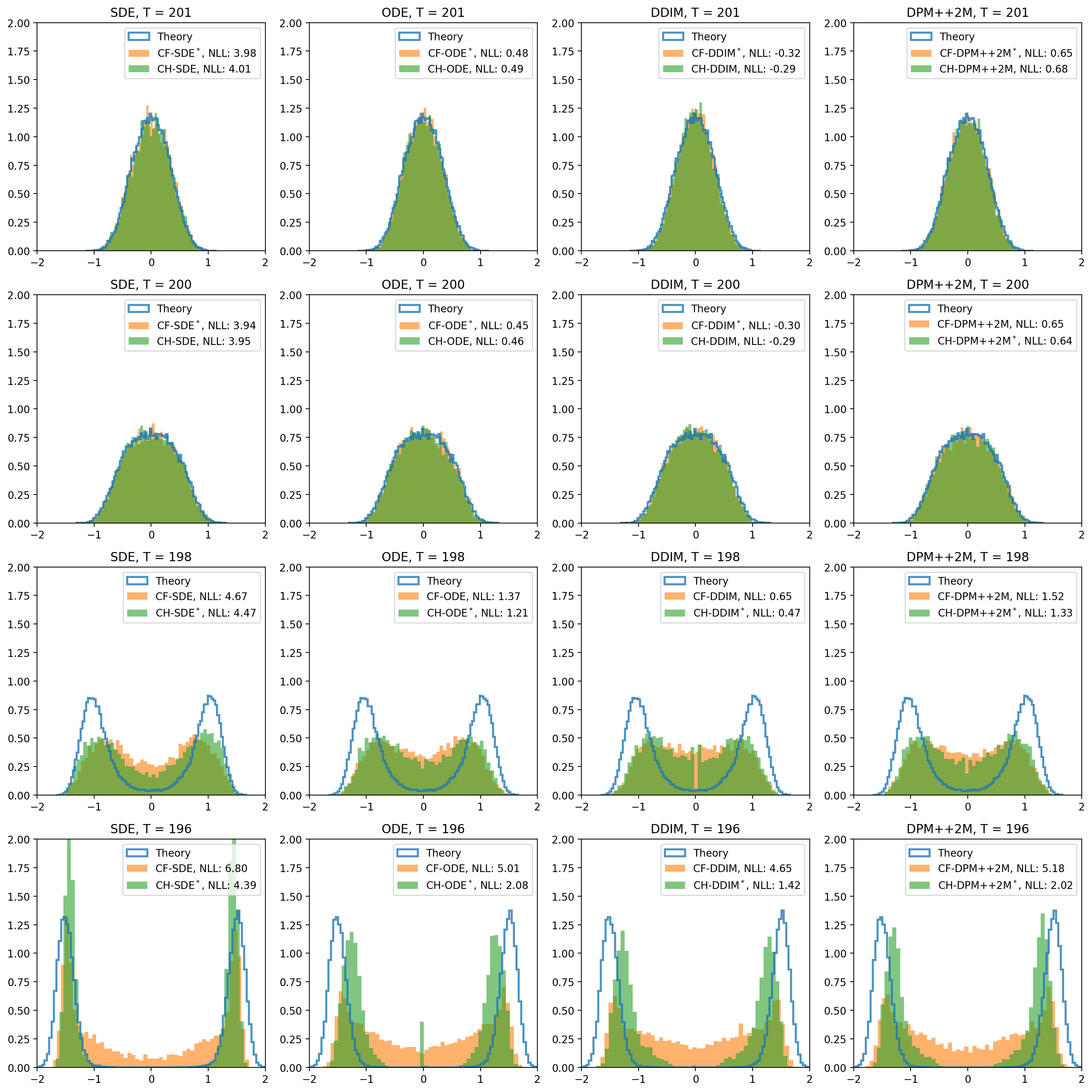}
\end{center}
\caption{The histogram of the mean magnetization of the Landau-Ginzburg magnet model around the Curie temperature $T=200$. The magnetization fields are generated by classifier-free and characteristic guidance. The characteristic guidance generates better samples below the Curie temperature $T=200$ and has better NLL. } \label{The sampling results phi4}
\end{figure}

We sample from guided DDPMs using four different ways: the SDE \eqref{1.5 backward diffusion process sampling discrete, alt}, probabilistic ODE \cite{Song2020ScoreBasedGM}, DDIM \cite{Song2020DenoisingDI}, and DPM++2M \cite{Lu2022DPMSolverFS}. During sampling, we set $\beta_1 = 1e-4$,
$\beta_2 = 0.015$ and the total time step to be $n=1000$ for SDE and ODE, and $n=20$ for DDIM and DPM++2M. For each temperature, we generate theoretical reference samples from the Boltzmann distribution in \eqref{Boltzmann distribution} with the Metropolis-Hastings algorithm. The negative log-likelihood (NLL) of samples of DDPM is computed as their mean Landau-Ginzburg Hamiltonian, subtracting the mean Landau-Ginzburg Hamiltonian of samples of the Metropolis-Hastings algorithm. The histogram of the mean value of the magnetization $\phi$ field values are plotted in Fig.\ref{The sampling results phi4}. 

A phase transition occurs at the Curie temperature $T_c=200$. Above the Curie temperature, the histogram of the mean magnetization has one peak centered at 0. At a certain instance, the thermal movements of molecules in our magnet may lead to a non-zero net magnetization, but the average magnetization over a long time is still zero, corresponding to a paramagnetic magnet. Below the Curie temperature, the histogram of the mean magnetization has two peaks with non-zero centers. Jumping from one peak to another at this case is difficult because the thermal movements of molecules only slightly change the mean magnetization and are insufficient to jump between peaks. This leads to a non-zero average magnetization over a long time and corresponds to a permanent magnet. Both classifier-free and characteristic guidance generate accurate samples above the Curie temperature where the DDPM is trained. However, the characteristic guidance generates better samples below the Curie temperature and has better NLL. Moreover, samples of characteristic guidance have well-separated peaks while samples of classifier-free guidance are not. This means the characteristic guidance has a better capability to model the phase transition. 

\section{Experimental Details on Cifar-10 and ImageNet 256}
 Our CIFAR-10 experiment utilized a modified U-Net model with 39 million parameters, sourced from a classifier-free diffusion guidance implementation \cite{classifierfreepytorch}. The model featured a channel size of 128 and a conditional embedding dimension of 10, operating over 1000 timesteps. Training involved 400,000 iterations with a batch size of 256, using the AdamW optimizer at a $2\times 10^{-4}$ learning rate and a 0.1 dropout rate. Stability was ensured with an EMA decay rate of 0.999, and the diffusion process followed a linear $\beta_t$ schedule from $10^{-4}$ to $0.02$ over 1000 timesteps. The characteristic guidance is computed by Alg.\ref{RMS ite} with $\gamma=0.002$, $\eta = 0.001$, and $\alpha = 0.999$. All CIFAR-10 experiments were conducted on 1 NVIDIA Geforce RTX 3090 GPU.

For the ImageNet 256 experiment, we utilized a pretrained latent diffusion model, as detailed in \cite{Rombach2021HighResolutionIS}, featuring over 400 million parameters. This model is configured to run with a total of 1000 timesteps and works on latents with the shape of $3\times 64\times 64$. The characteristic guidance is computed by Alg.\ref{AA ite} with $\gamma=1$, $m = 2$, $\eta = 0.005$. Experiments were conducted on 6 NVIDIA Geforce RTX 4090 GPUs.
  

\section{Visualization Results on Latent Diffusion Model}

\begin{figure*}[h]
\begin{center}
\includegraphics[width=0.8\columnwidth]{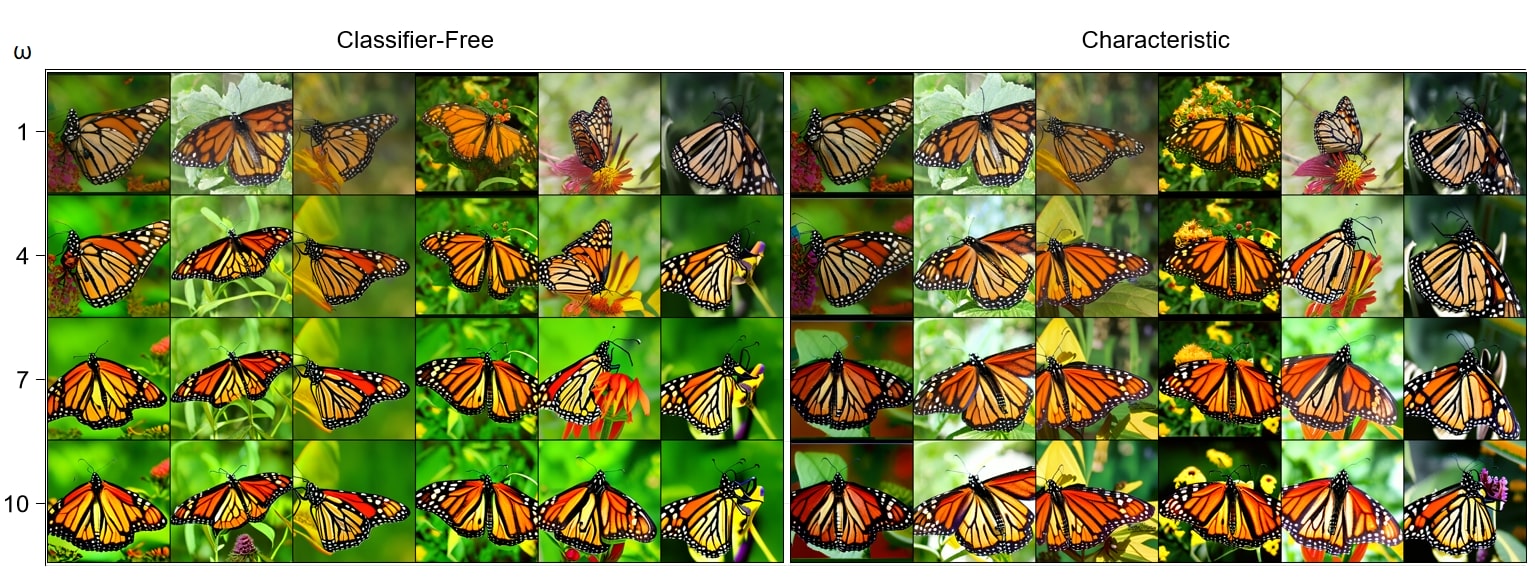}
\end{center}
\caption{Comparative visualization of butterfly (323) images generated from latent diffusion model using Classifier Free Guidance (CF) versus Characteristic Guidance (CH).} \label{Butterfly}
\end{figure*}
\begin{figure*}[h]
\begin{center}
\includegraphics[width=0.8\columnwidth]{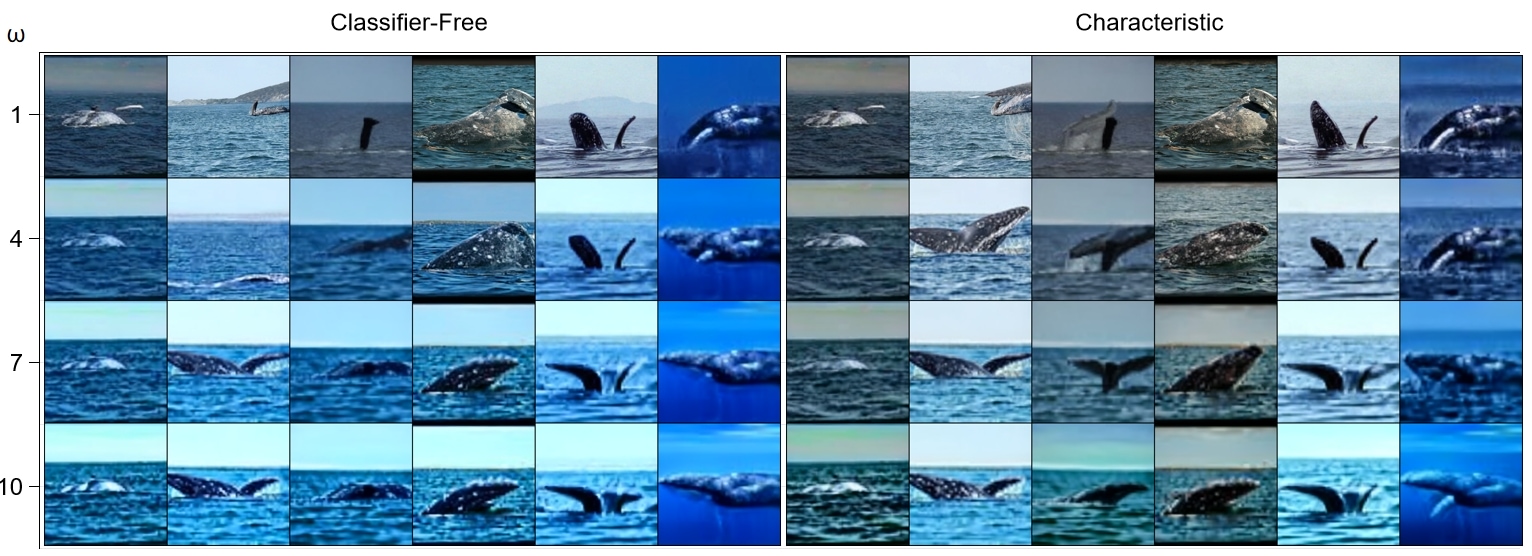}
\end{center}
\caption{Comparative visualization of whale (147) images generated from latent diffusion model using Classifier Free Guidance (CF) versus Characteristic Guidance (CH).} \label{Whale}
\end{figure*}

\begin{figure*}[h]
\begin{center}
\includegraphics[width=0.8\columnwidth]{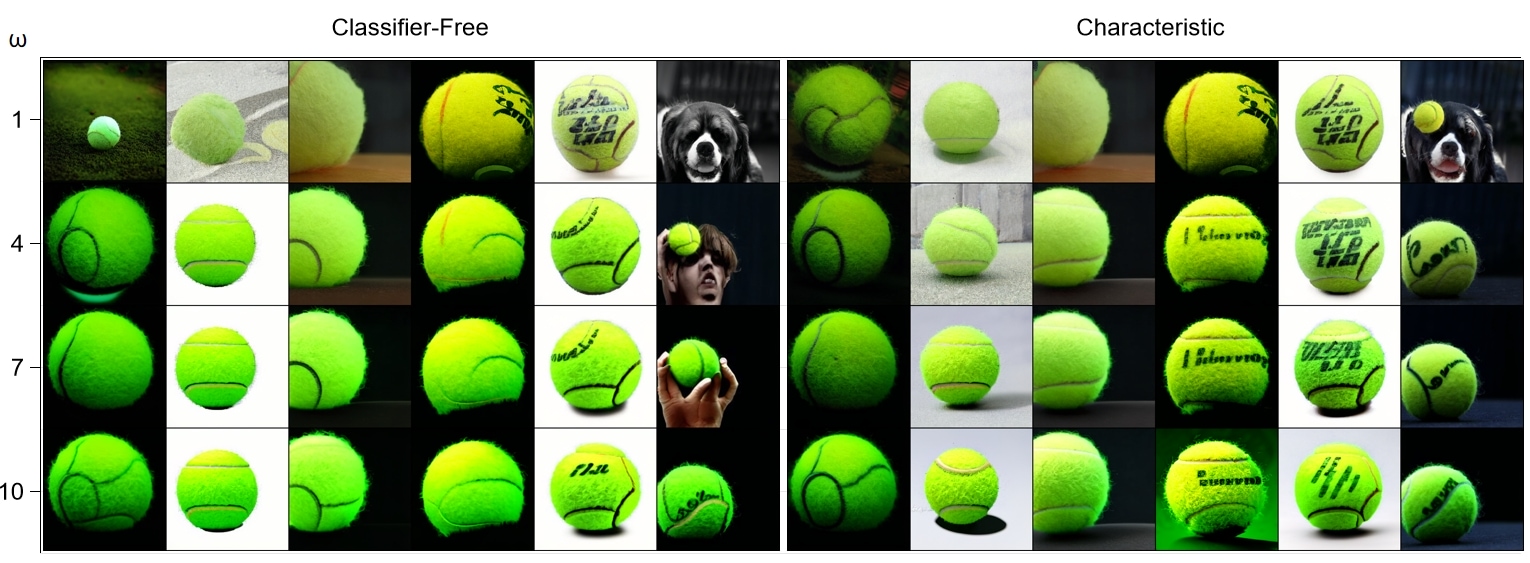}
\end{center}
\caption{Comparative visualization of butterfly (852) images generated from latent diffusion model using Classifier Free Guidance (CF) versus Characteristic Guidance (CH).} \label{Baseball}
\end{figure*}
\begin{figure*}[h]
\begin{center}
\includegraphics[width=0.8\columnwidth]{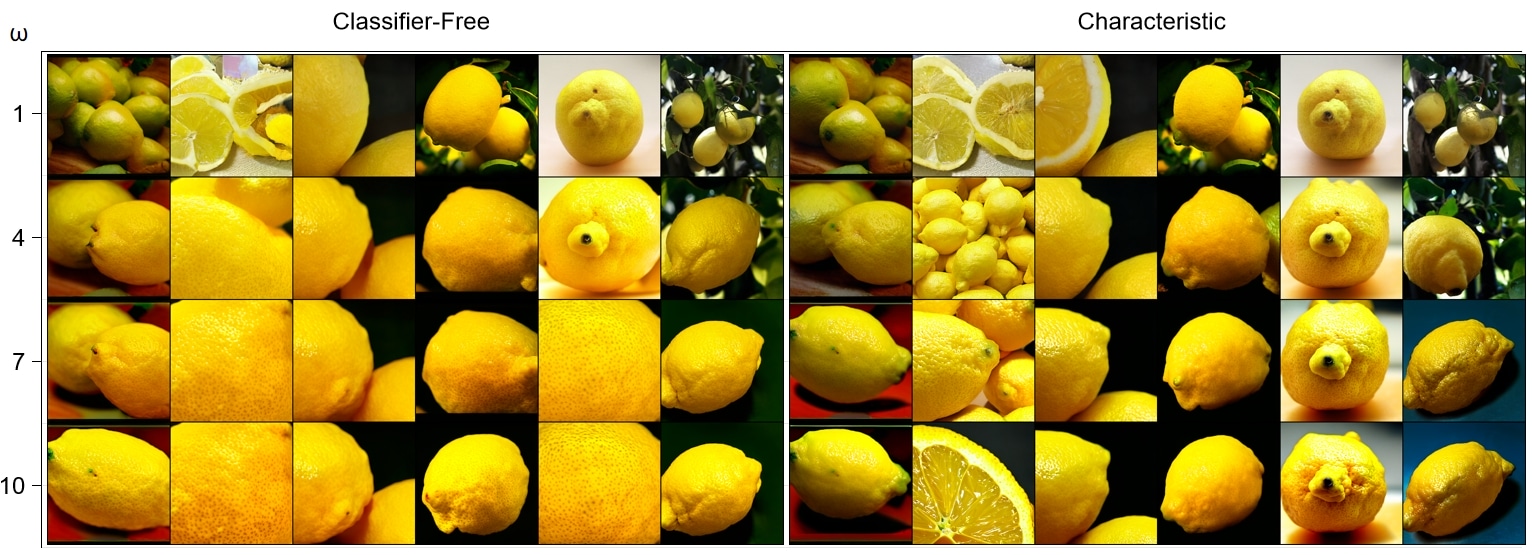}
\end{center}
\caption{Comparative visualization of lemon (147) images generated from latent diffusion model using Classifier Free Guidance (CF) versus Characteristic Guidance (CH).} \label{Lemon}
\end{figure*}

\begin{figure*}[h]
\begin{center}
\includegraphics[width=0.8\columnwidth]{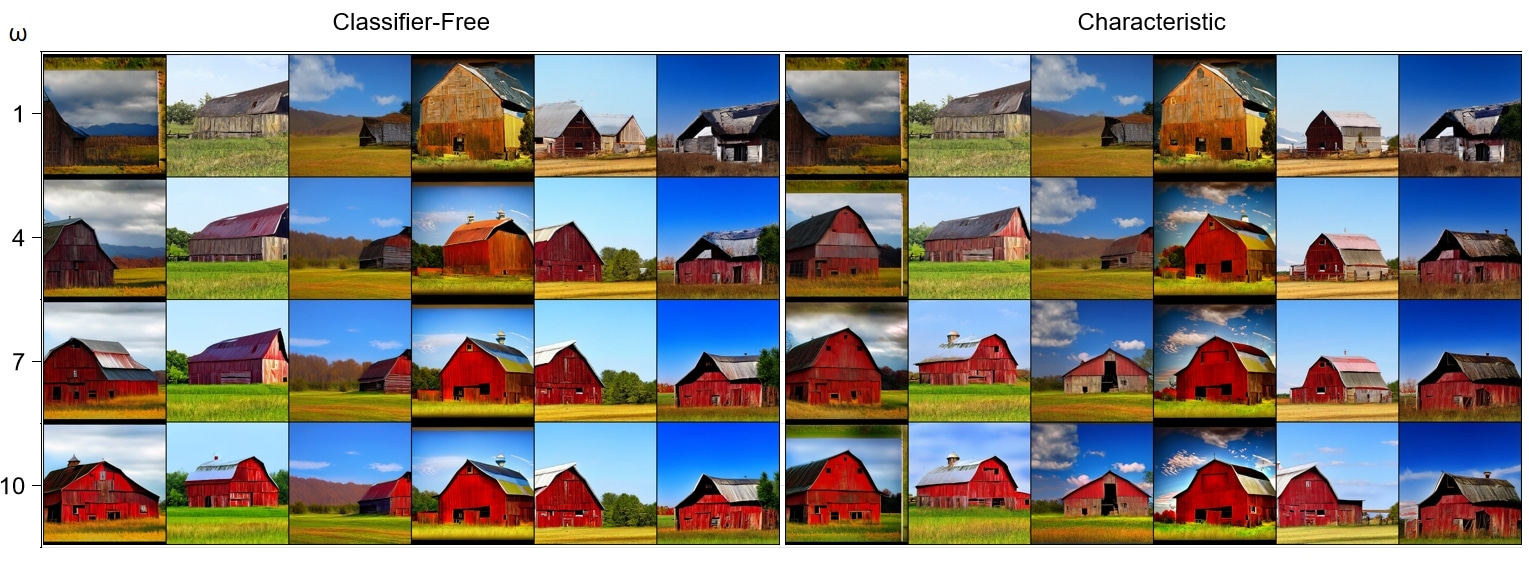}
\end{center}
\caption{Comparative visualization of barn (425) images generated from latent diffusion model using Classifier Free Guidance (CF) versus Characteristic Guidance (CH).} \label{Barn}
\end{figure*}

\begin{figure*}[h]
\begin{center}
\includegraphics[width=0.8\columnwidth]{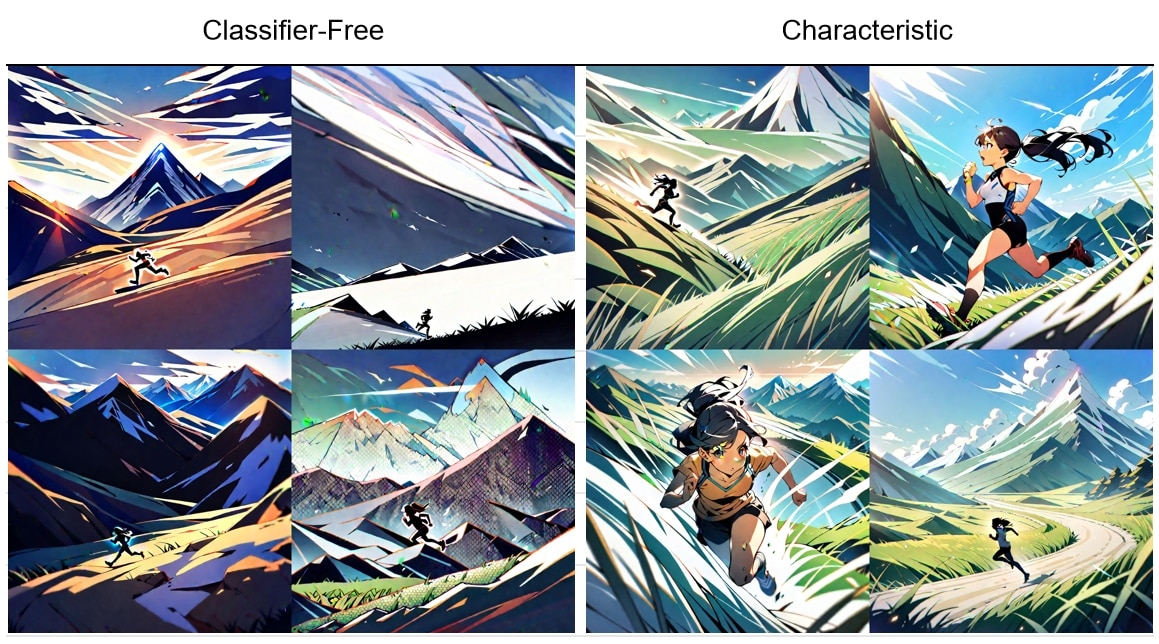}
\end{center}
\caption{1girl, running, mountain, grass. Comparative visualization of images generated from Stable Diffusion WebUI using Classifier Free Guidance versus Characteristic Guidance. Infotext of image (batch size 4, with the Characteristic Guidance Web UI extension): \textit{1girl, running, mountain, grass. Negative prompt: low quality, worst quality. Steps: 30, Sampler: UniPC, CFG scale: 30, Seed: 0, Size: 1024x1024, Model hash: 1449e5b0b9, Model: animagineXLV3\_v30, CHG: "\{RegS: 5, RegR: 1, MaxI: 50, NBasis: 1, Reuse: 0, Tol: -4, IteSS: 1, ASpeed: 0.4, AStrength: 0.5, AADim: 2, CMode: 'More ControlNet'\}", Version: v1.7.0}} \label{1grmg}
\end{figure*}

\begin{figure*}[h]
\begin{center}
\includegraphics[width=0.8\columnwidth]{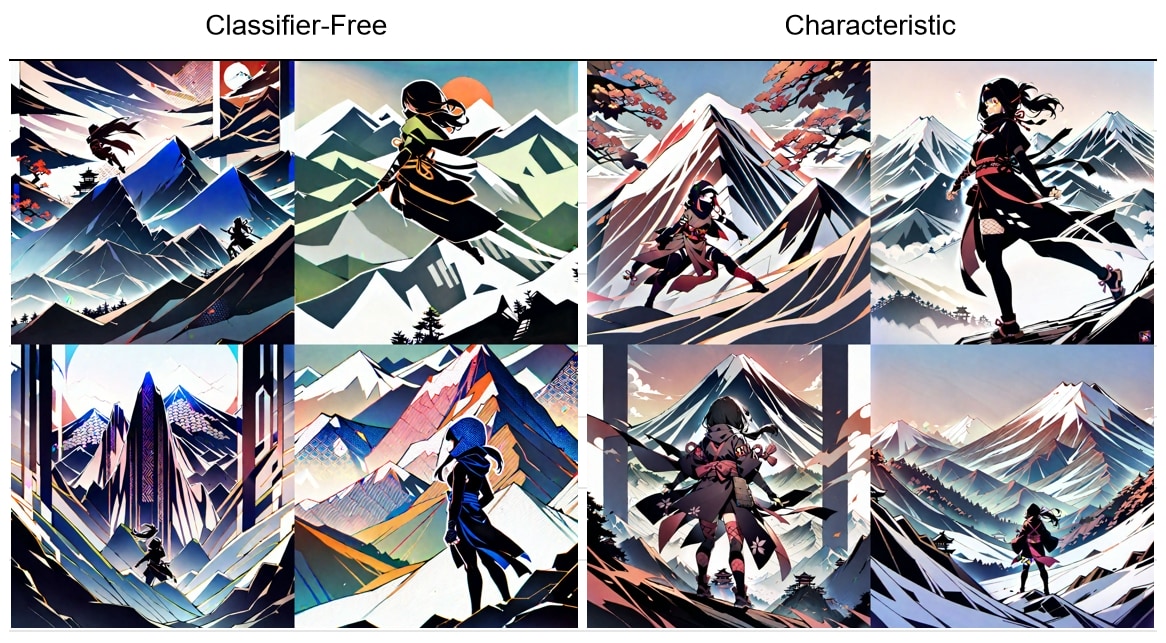}
\end{center}
\caption{1girl, ninja, mountain. Comparative visualization of images generated from Stable Diffusion WebUI using Classifier Free Guidance versus Characteristic Guidance. Infotext of image (batch size 4, with the Characteristic Guidance Web UI extension): \textit{1girl, ninja, mountain, Negative prompt: low quality, worst quality,
Steps: 30, Sampler: UniPC, CFG scale: 30, Seed: 0, Size: 1024x1024, Model hash: 1449e5b0b9, Model: animagineXLV3\_v30, CHG: "\{RegS: 5, RegR: 1, MaxI: 50, NBasis: 1, Reuse: 0, Tol: -4, IteSS: 1, ASpeed: 0.4, AStrength: 0.5, AADim: 2, CMode: 'More ControlNet'\}", Version: v1.7.0}} \label{1grmg}
\end{figure*}

\begin{figure*}[h]
\begin{center}
\includegraphics[width=0.8\columnwidth]{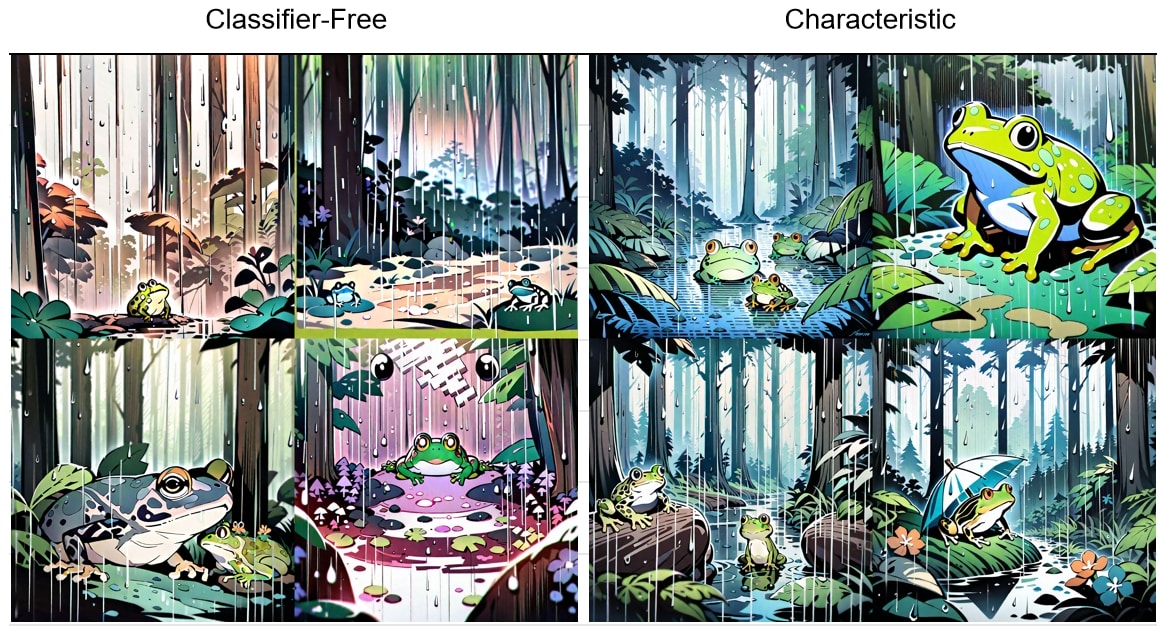}
\end{center}
\caption{frog, forest, rain. Comparative visualization of images generated from Stable Diffusion WebUI using Classifier Free Guidance versus Characteristic Guidance. Infotext of image (batch size 4, with the Characteristic Guidance Web UI extension): \textit{frog, forest, rain, Negative prompt: low quality, worst quality,
Steps: 30, Sampler: UniPC, CFG scale: 30, Seed: 0, Size: 1024x1024, Model hash: 1449e5b0b9, Model: animagineXLV3\_v30, CHG: "\{RegS: 5, RegR: 1, MaxI: 50, NBasis: 1, Reuse: 0, Tol: -4, IteSS: 1, ASpeed: 0.4, AStrength: 0.5, AADim: 2, CMode: 'More ControlNet'\}", Version: v1.7.0}} \label{1grmg}
\end{figure*}

\begin{figure*}[h]
\begin{center}
\includegraphics[width=0.8\columnwidth]{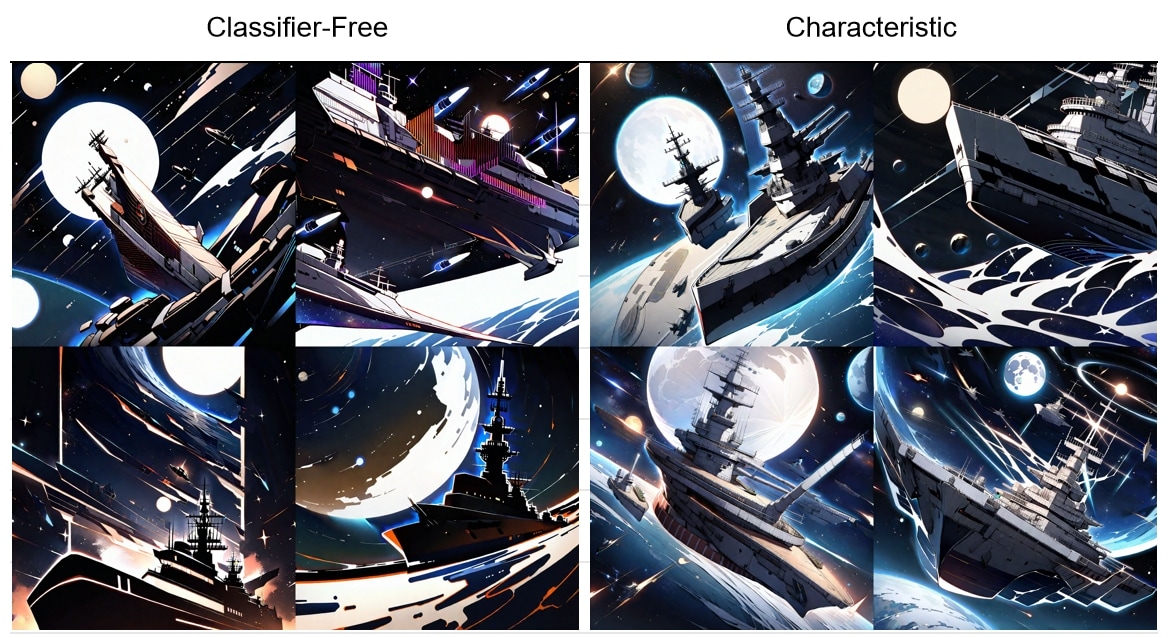}
\end{center}
\caption{battleship, space, moon. Comparative visualization of images generated from Stable Diffusion WebUI using Classifier Free Guidance versus Characteristic Guidance. Infotext of image (batch size 4, with the Characteristic Guidance Web UI extension): \textit{battleship, space, moon, Negative prompt: low quality, worst quality,
Steps: 30, Sampler: UniPC, CFG scale: 30, Seed: 0, Size: 1024x1024, Model hash: 1449e5b0b9, Model: animagineXLV3\_v30, CHG: "\{RegS: 5, RegR: 1, MaxI: 50, NBasis: 1, Reuse: 0, Tol: -4, IteSS: 1, ASpeed: 0.4, AStrength: 0.5, AADim: 2, CMode: 'More ControlNet'\}", Version: v1.7.0}} \label{1grmg}
\end{figure*}

\begin{figure*}[h]
\begin{center}
\includegraphics[width=\columnwidth]{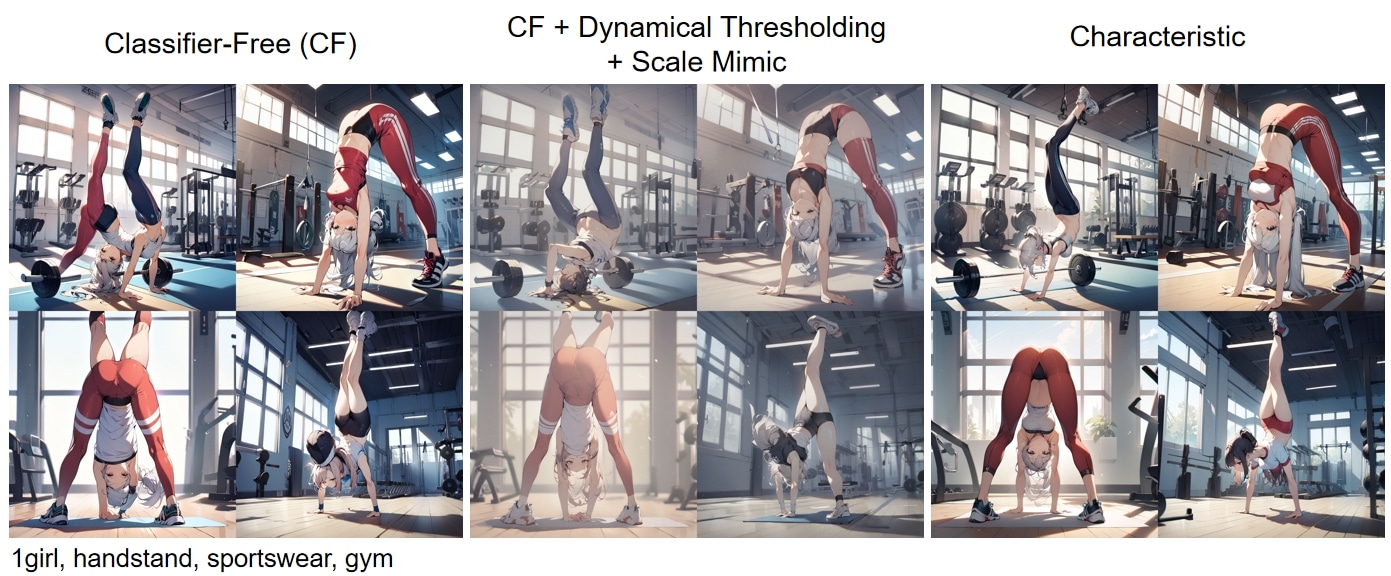}
\end{center}
\caption{1girl, handstand, sportswear, gym. Comparative visualization of images generated from Stable Diffusion WebUI using Classifier Free Guidance, Classifier Free Guidance + Dynamical thresholding + Scale Mimic, and Characteristic Guidance. The "scale mimic" technique is a workaround to suppress artifacts of dynamical thresholding for latent space generation tasks. The seeds used to generate these images are 0,1,2,3 indicating no cherry picking. \\ Infotext of image (batch size 4): \textit{1girl, handstand, sportswear, gym, Negative prompt: low quality, worst quality,
Steps: 30, Sampler: DPM++ 2M Karras, CFG scale: 10, Seed: 0, Size: 1024x1024, Model hash: 1449e5b0b9, Model: animagineXLV3\_v30, Version: v1.7.0} \\ Intotext of dynamic thresholding:
\textit{Dynamic thresholding enabled: True, Mimic scale: 7, Separate Feature Channels: True, Scaling Startpoint: MEAN, Variability Measure: AD, Interpolate Phi: 1, Threshold percentile: 100, Mimic mode: Half Cosine Down, Mimic scale minimum: 0,} \\ Intotext of characteristic guidance:
\textit{CHG: "\{RegS: 1, RegR: 1, MaxI: 50, NBasis: 1, Reuse: 0, Tol: -4, IteSS: 1, ASpeed: 0.4, AStrength: 0.5, AADim: 2, CMode: 'More ControlNet'\}", }
} \label{gymhandstand}
\end{figure*}

\end{document}